\documentclass[runningheads]{llncs}

\usepackage{eccv}

\usepackage{eccvabbrv}

\usepackage{graphicx}
\usepackage{booktabs}
\usepackage{makecell,tabularx}

\usepackage{pifont}
\usepackage{hyperref}

\newcommand{\myparagraph}[1]{\noindent \textbf{#1}}
\newcommand{\mysubparagraph}[1]{\noindent \textbf{\emph{#1}}}
\newcommand{\ownmethod}{VEON }
\newcommand{\cmark}{\ding{51}}
\newcommand{\xmark}{\ding{55}}

\begin{document}

% ---------------------------------------------------------------
% TODO REVIEW: Replace with your title
\title{VEON: Vocabulary-Enhanced Occupancy Prediction} 
% \title{VEON: Vocabulary-Enhanced Occupancy Prediction \\ 
%  --- Supplementary Material ---} 

% TODO REVIEW: If the paper title is too long for the running head, you can set
% an abbreviated paper title here. If not, comment out.
\titlerunning{VEON: Vocabulary-Enhanced Occupancy Prediction}

% TODO FINAL: Replace with your author list. 
% Include the authors' OCRID for the camera-ready version, if at all possible.
\author{Jilai Zheng\inst{1} \and
Pin Tang\inst{1} \and
Zhongdao Wang\inst{2} \and
Guoqing Wang\inst{1} \and \\
Xiangxuan Ren\inst{1} \and 
Bailan Feng\inst{2} \and
Chao Ma\inst{1}\thanks{Corresponding author.}}

% TODO FINAL: Replace with an abbreviated list of authors.
\authorrunning{J.~Zheng et al.}
% First names are abbreviated in the running head.
% If there are more than two authors, 'et al.' is used.

% TODO FINAL: Replace with your institution list.
\institute{MoE Key Lab of Artificial Intelligence, AI Institute, Shanghai Jiao Tong University \and Huawei Noah's Ark Lab \\ 
\email{\{zhengjilai,pin.tang,guoqing.wang,bunny\_renxiangxuan,chaoma\}@sjtu.edu.cn}
\email{\{wangzhongdao,fengbailan\}@huawei.com}}

\maketitle
\begin{abstract}

Perceiving the world as 3D occupancy supports embodied agents to avoid collision with any types of obstacle. 
While open-vocabulary image understanding has prospered recently, how to bind the predicted 3D occupancy grids with open-world semantics still remains under-explored due to limited open-world annotations. Hence, instead of building our model from scratch, we try to blend 2D foundation models, specifically a depth model MiDaS and a semantic model CLIP, to lift the semantics to 3D space, thus fulfilling 3D occupancy. 
However, building upon these foundation models is not trivial. 
First, the MiDaS faces the depth ambiguity problem, i.e., it only produces relative depth but fails to estimate bin depth for feature lifting. 
Second, the CLIP image features lack high-resolution pixel-level information, which limits the 3D occupancy accuracy. 
Third, open vocabulary is often trapped by the long-tail problem. 
To address these issues, we propose VEON for \textbf{V}ocabulary-\textbf{E}nhanced \textbf{O}ccupancy predictio\textbf{N} by not only assembling but also adapting these foundation models. 
We first equip MiDaS with a Zoedepth head and low-rank adaptation (LoRA) for relative-metric-bin depth transformation while reserving beneficial depth prior. Then, a lightweight side adaptor network is attached to the CLIP vision encoder to generate high-resolution features for fine-grained 3D occupancy prediction. Moreover, we design a class reweighting strategy to give priority to the tail classes. 
With only $46$M trainable parameters and zero manual semantic labels, VEON achieves $15.14$ mIoU on Occ3D-nuScenes, and shows the capability of recognizing objects with open-vocabulary categories, meaning that our VEON is label-efficient, parameter-efficient, and precise enough. 


\end{abstract}    
\section{Introduction}
\label{sec:intro}
In recent years, the autonomous driving community has been paying increasing attention to the sophisticated, voxel-level understanding of the 3D space around the ego car. 
This perception task of the new era, dubbed as \emph{Occupancy Prediction}, aims to assign each voxel in the 3D space with semantic information, namely what (class of) object occupies each specific voxel.
In this paper, we mainly focus on vision-centric open-vocabulary occupancy prediction. This practical setting stands out for \textit{(1)} utilizing only surrounding images during inference and \textit{(2)} recognizing objects of a variety of categories that could exist on the roads. Such geometrical and fine-grained information has been proven beneficial to not only the scene understanding but also the subsequent planning and control~\cite{cvpr23-uniad,iccv23-vad}. 

\begin{figure*}[t]
	\centering
	\includegraphics[width=0.85\linewidth,keepaspectratio]{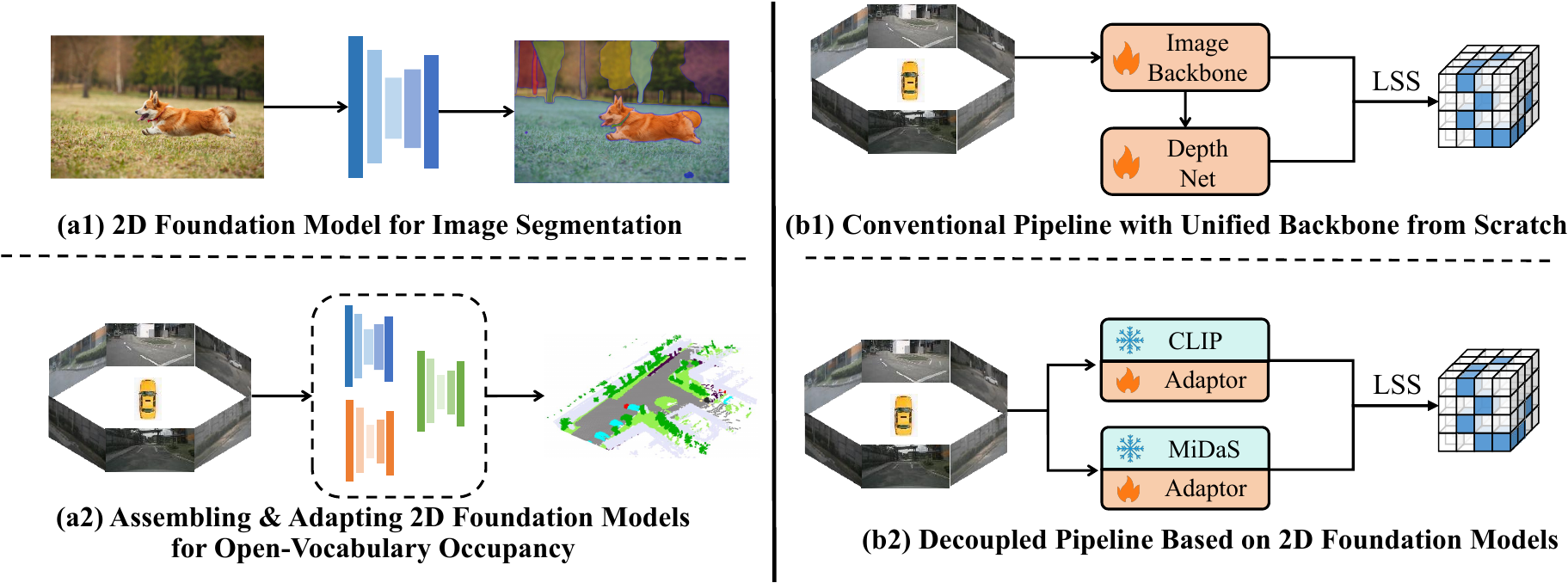}
		% \vspace{-5pt}
	\caption{Main idea of our VEON. \textbf{Left:} Referring to the strong data prior in 2D foundation models, we resort to unleashing their power for handling 3D open-vocabulary tasks. \textbf{Right:} Compared with the conventional practice of training a unified 2D backbone from scratch, we design a decoupled pipeline that assembles and adapts a depth model MiDaS~\cite{tpami20-midas} and a semantic model CLIP~\cite{icml21-clip}, for 3D open-vocabulary occupancy. } 
	\label{fig:idea}
		% \vspace{-10pt}
\end{figure*}

While there has been a remarkable improvement in open-vocabulary image understanding benefiting from 2D foundation models~\cite{icml21-clip} as shown in Fig.~\ref{fig:idea}(a1), their 3D counterparts on occupancy prediction still lag far behind. This is mainly attributed to the lack of large-scale open-world 3D occupancy annotations, which is caused by the labor-intensive labeling process. 
In fact, as illustrated in Fig.~\ref{fig:idea}(b1), existing solutions to open-vocabulary 3D occupancy prediction~\cite{arxiv23-ovo,nips23-pop3d,arxiv23-selfocc,arxiv23-occnerf} still rely on training an end-to-end deep network with depth estimation and semantic extraction modules from scratch. Considering the absence of abundant labeled open-vocabulary 3D data, the current strategy hinders the performance ceiling of open-vocabulary 3D occupancy predictors.

Inspired by the success of 2D open-vocabulary scene understanding, we alternatively resort to assembling 2D foundation models for open-world 3D occupancy and unleashing their power on 3D occupancy prediction as depicted in Fig.~\ref{fig:idea}(a2). 
A naive pipeline characterized by module decoupling is given in Fig.~\ref{fig:idea}(b2), where we utilize a depth foundation model MiDaS~\cite{tpami20-midas} to lift the semantics produced by the vision-language foundation model CLIP~\cite{icml21-clip} from 2D image pixels to 3D grids, thus fulfilling the 3D occupancy task. 
However, leveraging these foundation models is not trivial and meets challenges. First, as MiDaS is trained to estimate relative depth consistent across tens of indoor and outdoor datasets, a domain gap exists between the pretrained relative depth and the bin depth required in feature lifting~\cite{eccv20-lss}. Thus, we propose to first adapt MiDaS~\cite{tpami20-midas} with a Zoedepth~\cite{arxiv23-zoedepth} head for relative-to-metric depth transformation, and then convert metric depth to bin depth in a differentiable manner. Besides, we equip the MiDaS backbone with low-rank adaptation (LoRA)~\cite{iclr21-lora} to conduct domain transfer while reserving beneficial depth prior. Second, as CLIP~\cite{icml21-clip} is trained through image-level paired consistency, the CLIP image features lack spatial pixel-level information. Also, the ViT~\cite{arxiv20-vit} architecture causes a low-resolution compromise on the sizes of image features, which is fatal to scene understanding. To resolve this issue, we propose to attach a High-resolution Side Adaptor (HSA) beside the CLIP image encoder. It maintains high-resolution features to compensate for the information loss caused by the low-resolution CLIP encoder, and keeps lightweight by absorbing CLIP features. It can also slightly manipulate the CLIP attention bias, in order to make CLIP better suited to the requirement of spatial feature extraction. Finally, we also design a class reweighting loss to handle tail classes. By putting more emphasis on tail classes, our VEON could better learn to recognize various objects, sticking to the open-vocabulary essence.

Compared with the previous occupancy prediction methods, our VEON takes full advantage of the pretrained 2D foundation models with strong 2D data prior. It has much fewer trainable parameters while obtaining competitive performance. 
For example, with only $46.0$M trainable parameters and no manual semantic annotations, our VEON model (with ViT-L backbone) achieves a competitive performance of $15.14$ mIoU on the large-scale dataset Occ3D-nuScenes~\cite{cvpr20-nuscenes,arxiv23-occ3d}. It also demonstrates the capability of recognizing objects of unseen classes never explicitly annotated in the training dataset. 

Our main contributions can be summarized as follows.
\begin{itemize}
	\item We design a VEON framework to solve open-vocabulary 3D occupancy prediction by assembling and adapting a depth estimation foundation model (i.e.,  MiDaS~\cite{tpami20-midas}) and a vision-language foundation model (i.e., CLIP~\cite{icml21-clip}).  
	
	\item We propose to conquer the domain gap of applying MiDaS to occupancy prediction by relative-metric-bin transformation and low-rank adaptation. 
	
	\item We attach a lightweight side adaptor network beside CLIP for extracting high-resolution and spatial-aware features that better suit scene understanding. And a class reweighting loss is designed to put emphasis on tail classes. 
	
	\item Experiments show that our VEON can obtain competitive performance with very few trainable parameters and partial or even zero manual annotations.

\end{itemize}

\section{Related Work}
\label{sec:related}

\myparagraph{Vision-centric 3D occupancy prediction. } Occupancy prediction aims at assigning semantic labels to all voxels around the ego car~\cite{arxiv23-occ3d,arxiv24-occgen,cvpr24-sparseocc}. 
MonoScene~\cite{cvpr22-monoscene} is the first work on predicting voxel-wise occupancy given monocular RGB camera inputs. OccDepth~\cite{arxiv23-occdepth} exploits the stereo images and distills knowledge from them. TPVFormer~\cite{cvpr23-tpvformer} seeks a tri-perspective view representation to understand the scene. VoxFormer~\cite{cvpr23-voxformer} designs a lightweight framework for occupancy prediction by explicitly specifying visible voxel queries. While early works typically experiment on the SemanticKitti dataset~\cite{iccv19-semantickitti}, recently, several occupancy benchmarks have been built on larger-scale datasets. 
For instance, Occ3D~\cite{arxiv23-occ3d} explores a three-stage label generation pipeline for dense semantic occupancy labels. Annotations are generated on nuScenes~\cite{cvpr20-nuscenes} and Waymo~\cite{cvpr20-waymo}, and a novel CTF-Occ method is testified. 
Similarly, SurroundOcc~\cite{iccv23-surroundocc}, OpenOccupancy~\cite{iccv23-openoccpancy} and OccNet~\cite{iccv23-occnet} also constructs their occupancy benchmarks on nuScenes~\cite{cvpr20-nuscenes}.

% SurroundOcc~\cite{iccv23-surroundocc} processes the static scenes and moving objects separately, in order to generate dense occupancy labels on nuScenes~\cite{cvpr20-nuscenes}. 
% OpenOccupancy~\cite{iccv23-openoccpancy} is a nuScenes occupancy benchmark of high-resolution ground truth with human purification. 
% OccNet~\cite{iccv23-occnet} proposes an occupancy reconstruction method via a cascade and temporal voxel decoder. It focuses on subsequent tasks after obtaining occupancy, \eg motion planning. 

\myparagraph{Open-vocabulary 3D scene understanding.} Foundation 2D vision-language models establish a strong connection between natural language and images. However, this connection is lacking in 3D scene understanding. One natural solution is to connect 3D data and language by utilizing 2D as a bridge. 3D-OVS~\cite{nips23-3dovs} distills knowledge from CLIP~\cite{icml21-clip} and DINO~\cite{iccv21-dino} into a neural radiance field (NeRF~\cite{commacm21-nerf}), obtaining the capability of 3D open-vocabulary segmentation. PLA~\cite{cvpr23-pla} leverages the geometric consistency between posed images and 3D scenes to learn language-driven 3D representation. 
% OpenMask3D~\cite{nips23-openmask3d} first generates class-agnostic 3D mask proposals and then aligns per-mask embeddings with multi-view CLIP features.
OpenScene~\cite{cvpr23-openscene} predicts dense 3D scene representation via aligning the point features with CLIP.
OVIR-3D~\cite{icrl23-ovir3d} explores open-vocabulary 3D instance retrieval by first generating 2D text-aligned region proposals and then fusing them in 3D. 
OpenIns3D~\cite{arxiv23-OpenIns3D} proposes ``Mask-Snap-Lookup'' for open-vocabulary 3D instance segmentation.

\myparagraph{Open-vocabulary 3D occupancy prediction.} Predicting open-vocabulary occupancy remains an under-explored problem. Early works~\cite{arxiv23-ovo} mainly focus on small-scale scenes. Recently, POP-3D~\cite{nips23-pop3d} introduced this task into the nuScenes dataset~\cite{cvpr20-nuscenes} 
for autonomous driving. POP-3D is trained from scratch with the conventional 2D-3D encoder architecture, and leverages language, point cloud, and images for training. Some self-supervised occupancy predictors, \eg SelfOcc~\cite{arxiv23-selfocc} and OccNeRF~\cite{arxiv23-occnerf}, can also be revised for open-vocabulary recognition by aligning with pseudo open-vocabulary labels in 2D. 
% However, the above methods do not unleash the power of 2D foundation models for occupancy prediction, and lag far behind our VEON in terms of performance. 

\section{Method}
\label{sec:method}

\subsection{Problem Setup}
\label{sec:preliminary}

3D occupancy prediction focuses on predicting the voxel-wise semantic state in 3D space. We divide the space around the ego car into $H \times W \times Z$ voxels and predict which class of object occupies each voxel, denoted as $\mathbf{O}$. Here $H, W, Z$ are respectively the length, width, and height of the equally sliced voxel grid. During inference, the model will only input $N_{cam}$ images $\mathbf{I} = \left\{ I_{i} \mid i \in \left[ 1, N_{cam} \right] \right\}$ from surrounding cameras, implying a vision-centric occupancy prediction task. While in training, the corresponding point cloud $\mathbf{P}$ is also available. Notice that $\mathbf{P}$ is collected via LiDAR, without manual efforts.  

Our model is designed to recognize objects in an open-vocabulary setting. Formally, the overall class set $C_{all}$ can be divided into a seen class set $C_{s}$ and an unseen class set $C_{u}$, where $C_{s} \cup C_{u} = C_{all}$ and $C_{s} \cap C_{u} = \varnothing$. During training, our model needs to fit its prediction $\mathbf{O}$ to the ground truth $\mathbf{\hat{O}}$, where all the labeled classes in $\mathbf{\hat{O}}$ are inside $C_{s}$. During inference, the model is required to provide open-set occupancy results (inside $C_{all}$). 
Our work focuses on two settings, i.e., $C_{s} \not= \varnothing$ and $C_{s} = \varnothing$. 
The former utilizes partial semantic labels, while the latter has no access to any semantic annotations. 

\subsection{Framework Overview}

\begin{figure*}[t]
	\centering
	\includegraphics[width=0.87\linewidth,keepaspectratio]{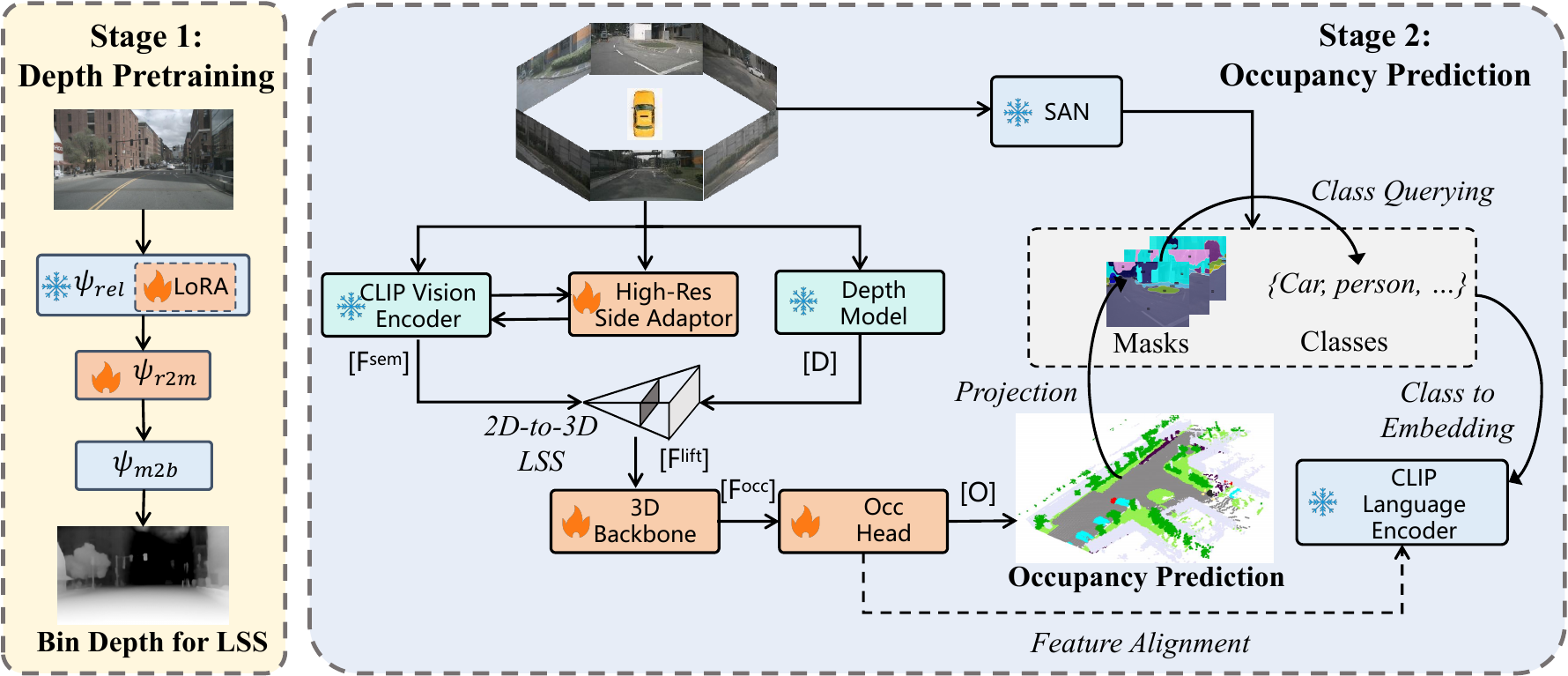}
	\caption{Framework overview. Our VEON consists of two training stages: depth pretraining and occupancy prediction. \textbf{Left:} In stage $1$, we adapt the MiDaS~\cite{tpami20-midas} backbone with a relative-metric-bin depth transformation adaptor to estimate the bin depth for LSS feature lifting~\cite{eccv20-lss}. Low-rank adaptation (LoRA)~\cite{iclr21-lora} is integrated for enhanced domain transfer. \textbf{Right:} In stage $2$, we unleash the power of CLIP~\cite{icml21-clip} via equipping a High-resolution Side Adaptor (HSA). The refined high-resolution CLIP semantic feature is lifted via LSS and goes through 3D convolutions for 3D occupancy. The network reserves the capability of recognizing open-vocabulary objects by aligning the 3D representation with CLIP language embeddings of certain classes, which is determined by the off-the-shelf 2D open-vocabulary segmentor SAN~\cite{cvpr23-san}. } 
	\label{fig:framework}
	% \vspace{-10pt}
\end{figure*}

Fig.~\ref{fig:framework} illustrates the basic framework of our VEON. The design rationale of VEON is to assemble and adapt two 2D foundation models, the large depth model MiDaS~\cite{tpami20-midas,arxiv23-midasv3} and the vision-language semantic-aware model CLIP~\cite{icml21-clip}, through a decoupled network structure. These two foundation models are trained with a vast number of 2D data, providing strong data prior for our VEON. 
As in Fig.~\ref{fig:framework}, we divide the training procedure of VEON into two stages as below:

\begin{itemize}
	
    \item \textbf{Stage 1: Depth Pretraining.} In stage 1, we adapt and tune a depth estimation model $\phi_{dp}$ from the foundation depth model MiDaS~\cite{tpami20-midas,arxiv23-midasv3}. $\phi_{dp}$ takes surrounding camera images $\mathbf{I}$ as input, and estimate bin depth $\mathbf{D}^{\prime}$ for them, ready for future LSS~\cite{eccv20-lss}. 

    \item \textbf{Stage 2: Occupancy Prediction.} In stage 2, we equip the CLIP~\cite{icml21-clip} vision encoder with a High-resolution Side Adaptor (HSA), in order to extract an enhanced semantic-aware 2D feature $\mathbf{F^{sem}}$. Then, we lift $\mathbf{F^{sem}}$ as $\mathbf{F^{lift}}$ via LSS~\cite{eccv20-lss} based on the bin depth $\mathbf{D}^{\prime}$ estimated through $\phi_{dp}$. Finally, we process $\mathbf{F^{lift}}$ via 3D convolutions, outputting the occupancy $\mathbf{O}$. During training, $\mathbf{O}$ will be projected and aligned with a 2D open-vocabulary segmentor. 
	
\end{itemize}

We note that leveraging these two foundation models is not easy due to several challenges, including domain gap, low resolution, tail classes, etc. Thus, as in Fig.~\ref{fig:framework}, we carefully design lightweight adaptors to unleash the power of these foundation models. We will go into particulars in the following sections. 

\subsection{Stage 1: Depth Pretraining}

MiDaS is a monocular depth estimation model trained with tens of labeled depth datasets~\cite{cvpr18-megadepth,cvpr20-structuredepth,eccv12-nyuv2,cvpr20-blendedmvs,tpami19-apolloscape}. 
To combine various depth datasets with distinct biases as a whole, MiDaS~\cite{tpami20-midas} establishes a solution by estimating relative depth irrelevant to depth range and scale. In this way, the pretrained MiDaS backbone, denoted as $\phi_{rel}$, could estimate precise relative depth across biased datasets. 

Despite the strong data prior provided by MiDaS, there exists a gap between the pretrained MiDaS and our requirements.  
In fact, MiDaS~\cite{tpami20-midas} is trained for relative depth, but LSS~\cite{eccv20-lss} in 3D occupancy requires normalized bin depth for 2D-to-3D view transformation. 
% but only bin depth is compatible with LSS~\cite{eccv20-lss}. 
Besides, the depth domain in autonomous driving is slightly different from that of the pretraining datasets. 
This motivates us to design the relative-metric-bin adaptor for end-to-end differentiable depth transformation. We also adopt the low-rank adaptation (LoRA)~\cite{iclr21-lora} to tune the MiDaS backbone for enhanced domain transfer. 

\myparagraph{Pipeline.} We propose to attach a \emph{relative-metric-bin adaptor} to the MiDaS backbone to transform the relative depth into bin depth for LSS and bridge the domain gap. 
As shown in Fig.~\ref{fig:framework} (left), we can formulate the depth estimation module as $\phi_{dp} = \phi_{rel} \circ \phi_{r2m} \circ \phi_{m2b}$. And the bin depth can be estimated by $\mathbf{D}^{\prime}=\phi_{dp}(\mathbf{I})$.
Here, $\circ$ means the cascade operation of networks. $\phi_{rel}$, $\phi_{r2m}$ and $\phi_{m2b}$ are respectively the relative depth backbone, the relative-to-metric adapting network, and the metric-to-bin transformation, as presented below.

\mysubparagraph{(1) Relative depth backbone $\phi_{rel}$.} We directly adopt the pretrained MiDaS to serve as $\phi_{rel}$, and freeze it for reserving the data prior. However, as there exists a domain gap between the pretraining data and driving scenes, we apply {low-rank adaptation (LoRA)}~\cite{iclr21-lora} to all linear layers within the MiDaS backbone. 
Notably, this strategy adds only $0.3\%$ additional parameters to the pretrained MiDaS, but significantly enhances domain transfer and unleashes the power of the depth foundation model as shown in Sec.~\ref{sec:ablation}. 

\mysubparagraph{(2) Relative-to-metric adapting network $\phi_{r2m}$.} Metric depth represents depth with absolute values (\eg 50 meters). For building $\phi_{r2m}$, we introduce the ZoeDepth~\cite{arxiv23-zoedepth} head as a lightweight network adaptor. This module collects features from decoder layers of the MiDaS backbone and leverages an enhanced bin-based strategy~\cite{eccv22-localbins,cvpr21-adabins} for calculating the metric depth. We refer readers to the ZoeDepth~\cite{arxiv23-zoedepth} paper for detailed network architecture. 

We optimize $\phi_{r2m}$ by fitting the metric depth $\mathbf{D}$ output from $\phi_{r2m}$ towards the ground truth depth $\mathbf{\hat{D}}$. 
Here $\mathbf{\hat{D}}$ is obtained by projecting the point cloud $\textbf{P}$ onto the camera plane. Suppose $d_{i}$ is the $i$-th pixel of $\mathbf{D}$, and $\hat{d}_{i}$ is the corresponding ground truth. We strictly follow \cite{nips14-depth,arxiv23-zoedepth,cvpr21-adabins} to formulate a pixel-wise scale-invariant depth loss $L_{pix}$ (see the supplementary material for formulation).
$L_{pix}$ ensures the shape and smoothness of the output metric depth, beneficial to the subsequent bin depth transformation. 

\mysubparagraph{(3) Metric-to-bin transformation $\phi_{m2b}$.} To transform metric depth $\mathbf{D}$ to bin depth $\mathbf{D}^{\prime}$ for LSS, we define $N_{bin}$ depth bins with equal width $w$. Suppose the first depth bin has its center as $d_{fc}$, then the $j^{th}$ depth bin ($0 \le j < N_{bin}$) should cover the interval $\left[d_{fc} + (j - 0.5) \cdot w, d_{fc} + (j + 0.5) \cdot w\right]$ with bin center $d_{fc} + j \cdot w$. Then, the metric depth $d_{i}$ can be transformed into a $N_{bin}$ dimension tensor $d_{i}^{\prime}$ (bin depth), with the $j^{th}$ dimension representing the similarity score of $d_{i}$ to the $j^{th}$ depth bin. We formally define this similarity value $d^{\prime}_{ij}$ as:
\begin{equation}\label{eq:bin_simialrity}
	d^{\prime}_{ij} =  \operatorname{softmax}_{j} (\beta \cdot h_{ij}), \ \ \text{where} \ \  h_{ij} = - \left| d_{i} - d_{fc} - j \cdot w \right|,
\end{equation}
and $\beta$ is a constant. Then, we can define the ground truth one-hot depth bin distribution $\hat{d}^{\prime}_{ij}$ as follows: 
\begin{equation}\label{eq:bin_gt}
\hat{d}^{\prime}_{ij} = \left\{
	\begin{array}{ll}
	1,   & \quad \text{if} \ \ | \hat{d}_{i} - d_{fc} - j \cdot w | \le w / 2 \\
	0,   & \quad \text{otherwise}
	\end{array}\right.
\end{equation}
In this way, the bin depth map $\mathbf{D^{\prime}}$ can be supervised with cross-entropy loss, with the loss defined as $L_{bd}$. The total loss in stage one $L_{stg1}$ is the weighted sum of the pixel-wise depth loss $L_{pix}$ and the bin depth loss $L_{bd}$.

\subsection{Stage 2: Occupancy Prediction} In stage $2$, we resort to CLIP for extracting a 2D semantic-aware feature $\mathbf{F^{sem}}$, and then lift $\mathbf{F^{sem}}$ from 2D to 3D via LSS~\cite{eccv20-lss} based on the bin depth map $\mathbf{D^{\prime}}$ from Eq.~\ref{eq:bin_simialrity}. 
A trivial solution here is to directly fetch visual tokens from CLIP as $\mathbf{F^{sem}}$, but this meets challenges and we will discuss our improvement later. 
The LSS operation gives the initial 3D feature $\mathbf{F^{lift}}$. After that, the lifted feature $\mathbf{F^{lift}}$ will be processed via a series of ResNet3D~\cite{cvpr16-resnet} blocks, generating a dense semantic-aware representation, denoted as $\mathbf{F^{occ}}$. 

Then, we decode the occupancy results from $\mathbf{F^{occ}}$ through two separate 3D convolution heads. For each voxel-wise occupancy representation $F_{i}^{occ}$ from the $i^{th}$ voxel of $\mathbf{F^{occ}}$, we respectively adopt: \textit{(1)} a two-layer 3D convolution head to generate a tag $O_{i}^{bin} \in [0, 1]$ indicating binary occupancy state, i.e., whether the voxel is occupied by any object or not, and \textit{(2)} a three-layer 3D convolution head to predict a semantic-aware embedding $O_{i}^{sa}$ fitting the feature distribution of CLIP output. The embedding map $\mathbf{O^{sa}}$ above is responsible for determining the semantic class. 
During inference, suppose the $j^{th}$ class in any class set $C$ has its embedding from CLIP language encoder as $F_{j}^{lan}$, then the final occupancy result $O_{i}$ for the $i^{th}$ voxel can be formulated as follows:
\begin{equation}\label{eq:occ_output}
O_{i} = \left\{
\begin{array}{ll}
\operatorname{argmax}_{j} O_{i}^{sa} \cdot F_{j}^{lan},   & \quad \text{if} \ \ O_{i}^{bin} \ge \tau \\
0,   & \quad  otherwise
\end{array}\right. ,
\end{equation}
where class $0$ indicates the special class ``free'' and $0 < \tau < 1$. Here, we have $C = C_{all}$ for open-vocabulary occupancy prediction. In this way, we can assemble the 2D foundation models to formulate the 3D occupancy pipeline.

However, integrating CLIP meets two challenges. First, the resolution of CLIP features is small ($16 \times 44$ for ViT-B, and $18 \times 50$ for ViT-L), hindering fine-grained scene understanding. We thus maintain an adaptor network beside the CLIP vision encoder, reserving high-resolution information. Second, the CLIP tokens focus more on image-level information than spatial information, limiting perception performance. We thus propose slightly manipulating the feature extraction process by adding attention bias to transformer layers inside CLIP. 

\begin{figure*}[t]
	\centering
	\includegraphics[width=0.87\linewidth,keepaspectratio]{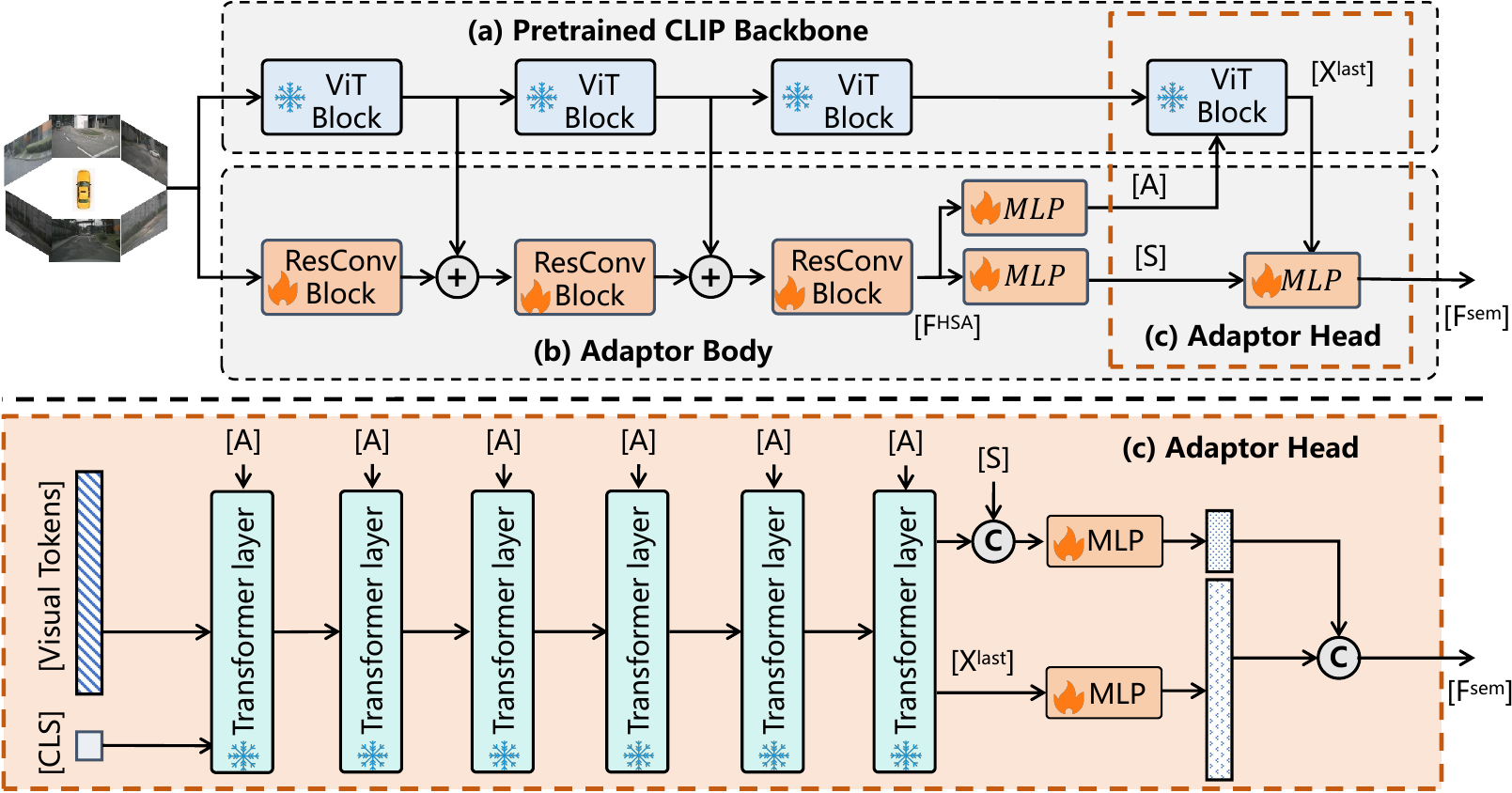}
	\caption{Detailed network architecture of the High-resolution Side Adaptor (HSA). \textbf{Top}: Adaptor architecture. We maintain a series of residual convolution blocks beside the CLIP backbone and extract high-resolution spatial features. It fuses early layers of the CLIP visual tokens and outputs: (1) attention bias ($\textbf{A}$) for refining ViT feature extraction, and (2) supplementary matrix ($\textbf{S}$) for making up high-resolution information. \textbf{Bottom}: Attention bias $\textbf{A}$ manipulates the attention of transformer layers in ViT, and $\textbf{S}$ is fused before outputting the 2D semantic feature $\mathbf{F^{sem}}$ for LSS lifting. } 
	\label{fig:detail}
		% \vspace{-20pt}
\end{figure*}

\myparagraph{High-resolution side adaptor (HSA).} As illustrated in Fig.~\ref{fig:detail}, our High-resolution Side Adaptor (HSA) can be divided into the adaptor body and the adaptor head. \emph{The adaptor body} consists of several residual blocks~\cite{cvpr16-resnet} parallel with the CLIP encoder and fuses multi-layer visual tokens from CLIP into the HSA body. Take the ViT-L CLIP variant with $24$ transformer layers as an example. We fuse visual tokens from the $6^{\text{th}}$ and $12^{\text{th}}$ CLIP layers into the features after the $1^{\text{st}}$ and $2^{\text{rd}}$ HSA body blocks. 
Since features in HSA have a higher resolution ($32 \times 88$) compared with visual tokens in CLIP ($18 \times 50$), 
we resize the CLIP visual tokens to be the same size as HSA features, and then fuse them with element-wise addition after channel alignment via $1 \times 1$ convolutions. 
% We note that more complex design may lead to better performance, but here we keep a clean and lightweight baseline. 
As shown in Fig.~\ref{fig:detail}, the HSA body accompanies the first $3/4$ layers ($18$ out of $24$ in ViT-L) of the CLIP backbone, resulting in a high-resolution feature map $\mathbf{F^{HSA}}$. 

On the other hand, \emph{the adaptor head} is responsible for manipulating the feature extraction process of the last $1/4$ layers of the CLIP backbone, making them more suitable for scene understanding. Specifically, we first apply two MLPs on $\mathbf{F^{HSA}}$ to obtain an attention bias matrix $\mathbf{A}$ and a supplementary matrix $\mathbf{S}$, as visualized in Fig.~\ref{fig:detail}. The first matrix $\mathbf{A}$ is the attention bias for CLIP visual tokens. Specifically, take the calculation of the attention process within the $i^{th}$ transformer layer in ViT as an example. We formulate this process as follows:
\begin{equation}\label{eq:attention_bias}
	\mathbf{X}_{i + 1} = \operatorname{softmax}(\mathbf{Q}_{i}\mathbf{K}_{i}^{T} + \mathbf{A}_{i} \mathbf{A}_{i}^{T}) \mathbf{V}_{i}.
\end{equation}
Here $\mathbf{X}_{i}$ represents the visual tokens in the $i^{th}$ layer, and  $\mathbf{Q}_{i}$, $\mathbf{K}_{i}$ and $\mathbf{V}_{i}$ are the linear transformations of $\mathbf{X}_{i}$. The attention bias $\mathbf{A}_{i}\mathbf{A}_{i}^{T}$ for the $i^{th}$ layer is added to $\mathbf{Q}_{i}\mathbf{K}_{i}^{T}$ for directing the transformer to pay more attention on the spatial information. 
Here, we neglect some elements (\eg, multiple attention heads) for convenience. Please refer to the supplementary material for details. 

We assemble the visual tokens $\mathbf{X^{last}}$ after the last transformer layer and the supplementary matrix $\mathbf{S}$ into the feature $\mathbf{F^{sem}}$ for LSS lifting. We first reshape and interpolate $\mathbf{X^{last}}$ to become the same shape as $\mathbf{S}$, and then construct $\mathbf{F^{sem}}$ via MLPs and concatenation:
\begin{equation}\label{eq:fsem}
\mathbf{F^{sem}} = \left[ \operatorname{MLP}_{1}(\mathbf{X^{last}}), \operatorname{MLP}_{2}(\left[\mathbf{X^{last}}, \mathbf{S} \right] )  \right]
\end{equation}
Here, square brackets denote feature concatenation. As in Fig.~\ref{fig:detail}, the output channel number of $\operatorname{MLP}_{1}$ is larger than that of $\operatorname{MLP}_{2}$ (i.e., $3:1$), as $\mathbf{S}$ is only designed as supplement to $\mathbf{X^{last}}$ provided by the CLIP encoder. 

\myparagraph{Training strategy.} Our VEON is optimized with joint supervision on $\mathbf{O^{bin}}$ and $\mathbf{O^{sa}}$. Specifically, for the binary occupancy state $\mathbf{O^{bin}}$, we adopt cross entropy (CE) to construct the binary occupancy loss $L_{bin}$. 
Its ground truth $\mathbf{\hat{O}^{bin}}$ can be derived from the point cloud $\mathbf{P}$ via offline post-processing~\cite{arxiv23-occ3d}. 
As for supervising the semantic-aware embedding map $\mathbf{O^{sa}}$, we enforce each embedding $O_{i}^{sa}$ of the $i^{th}$ voxel to match the (pseudo) ground truth CLIP class embedding for the $i^{th}$ voxel, namely $\hat{O}_{i}^{sa}$. The assignment of $\mathbf{\hat{O}^{sa}}$ is critical. Here we apply an off-the-shelf 2D open-vocabulary segmentor SAN~\cite{cvpr23-san} as the pseudo ground truth. In the case that $C_{s} = \varnothing$, we project the $i^{th}$ voxel onto the surrounding images based on the intrinsic and extrinsic camera parameters, and fetch the CLIP language embedding of corresponding open-vocabulary class $j$ (output from SAN~\cite{cvpr23-san}) as optimization target, namely $\hat{O}_{i}^{sa} = F_{j}^{lan}$. Otherwise, if $C_{s} \not= \varnothing$, $\hat{O}_{i}^{sa}$ is replaced with the ground truth class embedding if and only if the annotation exists. 

Then, we construct the feature alignment loss $L_{sa}$ via cosine similarity. In other words, the feature alignment loss for each voxel $i$ is calculated as $1 - \operatorname{cosine}(O_{i}^{sa}, \hat{O}_{i}^{sa})$. Traditionally, the cosine loss items of all voxels are averaged for calculating $L_{sa}$. However, as tail classes seldom exist in the training set, the vast majority of voxels will be trained to align with stuff classes (\eg, road, grass) in this case, which is harmful to open-vocabulary recognition. In this paper, we propose to reweight the loss component of each voxel as follows: 
\begin{equation}\label{eq:reweight}
L_{sa} = \frac{1}{\left|C\right|} \sum_{j \in C} \frac{1}{N_{j}^{\prime}} \sum_{\hat{O}_{i} = j}  1 - \operatorname{cosine}(O_{i}^{sa}, \hat{O}_{i}^{sa}).
\end{equation}
Here $N^{\prime}_{j} = \left| \left\{ i \mid \hat{O}_{i} = j \right\} \right|$, and $\hat{O}_{i}$ is calculated similar to Eq.~\ref{eq:occ_output} except that $O^{sa}_{i}$ is replaced with $\hat{O}^{sa}_{i}$. Eq.~\ref{eq:reweight} averages the voxel-level loss items within each class first, and then across all the classes. As tail classes occupy a much smaller number of voxels, they are prioritized during network optimization. Experiments prove that this design significantly alleviates the problem of tail classes. 
Finally, the loss in the training stage 2 $L_{stg2}$ is the weighted sum of $L_{bin}$ and $L_{sa}$.

\section{Experiments}
\label{sec:experiment}

\subsection{Experimental Settings}
\label{sec:dataset}

\myparagraph{Dataset.} 
Throughout our experiments, we select Occ3D~\cite{arxiv23-occ3d} for evaluating our proposed VEON. 
Occ3D~\cite{arxiv23-occ3d} is built on $700$ training scenes and $150$ validation scenes in the nuScenes dataset~\cite{cvpr20-nuscenes}. For each scene snapshot, nuScenes provides $6$ images from surrounding cameras, the camera parameters for view transformation, and LiDAR point clouds. 
Beyond that, Occ3D~\cite{arxiv23-occ3d} additionally annotates voxel-level semantic labels to serve as a benchmark for 3D occupancy prediction. 
These label masks have the resolution of $200 \times 200 \times 16$, with X-axis, Y-axis and Z-axis ranging respectively as $[-40, 40]$, $[-40, 40]$ and $[-1.0, 5.4]$ meters. The voxel size is $(0.4, 0.4, 0.4)$ meters. Following nuScenes LiDAR segmentation~\cite{ral22-panopticnuscenes}, one class out of $18$ classes (with a special class ``free'') is assigned to each voxel. We collect the IoUs on all $17$ normal classes (excluding ``free''), and a mean IoU (mIoU) of these classes as the evaluation metrics. 
Following the Occ3D-nuScenes protocol, we only consider visible voxels during evaluation. 

\myparagraph{Implementation.} We implement our VEON based on the BEVDet codebase~\cite{bevdet-codebase}. Our experimental settings follow BEVDet~\cite{arxiv21-bevdet,arxiv22-bevdet4d}, with the same data sampling, cropping, and augmentation strategy. We also employ the bevpoolv2~\cite{bevpoolv2} in BEVDet for fast LSS~\cite{eccv20-lss}. We select AdamW~\cite{iclr18-adamw} to be our network optimizer, with learning rate as $10^{-4}$ and weight decay as $10^{-2}$. All our experiments are performed on $8$ NVIDIA Tesla V100 GPUs, with a batch size of $1$ on each GPU. 
During training stage 1, we adopt the MiDaS~\cite{tpami20-midas} with BEiT-L backbone~\cite{iclr21-beit} as our depth foundation model $\phi_{rel}$ and initialize the weights by pretraining on a mixed set of $12$ depth datasets~\cite{arxiv23-zoedepth}. The camera input size for $\phi_{dp}$ is $256 \times 704$. 
During training stage 2, we load the 2D open-vocabulary semantic segmentor SAN~\cite{cvpr23-san} to generate pseudo labels for occupancy supervision. Since a frozen CLIP encoder exists inside SAN, we reuse the CLIP image encoder within our VEON framework. We test two variants of VEON throughout our experiments. VEON-B adopts the ViT-B CLIP variant with $12$ transformer layers for semantic extraction, while the larger VEON-L adopts the ViT-L CLIP with $24$ layers. 
The input image size in stage 2 is set as $512 \times 1408$.

\subsection{Main Results}
\label{sec:eff}

\definecolor{others}{RGB}{175, 0, 75}
\definecolor{barrier}{RGB}{255, 120, 50}
\definecolor{bicycle}{RGB}{255, 192, 203}
\definecolor{bus}{RGB}{255, 255, 0}
\definecolor{car}{RGB}{0, 255, 255}
\definecolor{const. veh.}{RGB}{255, 0, 255}
\definecolor{motorcycle}{RGB}{200, 180, 0}
\definecolor{pedestrian}{RGB}{255, 0, 0}
\definecolor{traffic cone}{RGB}{255, 240, 150}
\definecolor{trailer}{RGB}{135, 60, 0}
\definecolor{truck}{RGB}{160, 32, 240}
\definecolor{drive. suf.}{RGB}{139, 137, 137}
\definecolor{other flat}{RGB}{0, 150, 245}
\definecolor{sidewalk}{RGB}{75, 0, 75}
\definecolor{terrain}{RGB}{150, 240, 80}
\definecolor{manmade}{RGB}{230, 230, 250}
\definecolor{vegetation}{RGB}{0, 175, 0}

\begin{table*}[t]
	\setlength{\tabcolsep}{0.004\linewidth}
	\newcommand{\classfreq}[1]{{~\tiny(\semkitfreq{#1}\%)}}  %
	\centering
		\caption{Performance of our \ownmethod on Occ3D-nuScenes occupancy benchmark~\cite{cvpr20-nuscenes,arxiv23-occ3d} (validation set) with $C_{s} = \varnothing$. 
		We compare the VEON variants with existing occupancy predictors trained with (rows $1$-$6$) or without (rows $7$-$9$) manual labels. }
	\resizebox{1\linewidth}{!}{
	\begin{tabular}{l|c| c c c c c c c c c c c c c c c c c | c}
			
			\toprule
			Method
			& \rotatebox{90}{sem.}
			& \rotatebox{90}{\textcolor{others}{$\blacksquare$} oth.} 
			& \rotatebox{90}{\textcolor{barrier}{$\blacksquare$} bar.} 
			& \rotatebox{90}{\textcolor{bicycle}{$\blacksquare$} bic.}
			& \rotatebox{90}{\textcolor{bus}{$\blacksquare$} bus} 
			& \rotatebox{90}{\textcolor{car}{$\blacksquare$} car} 
			& \rotatebox{90}{\textcolor{const. veh.}{$\blacksquare$} c. v.} 
			& \rotatebox{90}{\textcolor{motorcycle}{$\blacksquare$} mot.} 
			& \rotatebox{90}{\textcolor{pedestrian}{$\blacksquare$} ped.} 
			& \rotatebox{90}{\textcolor{traffic cone}{$\blacksquare$} t. c.} 
			& \rotatebox{90}{\textcolor{trailer}{$\blacksquare$} tra.} 
			& \rotatebox{90}{\textcolor{truck}{$\blacksquare$} tru.} 
			& \rotatebox{90}{\textcolor{drive. suf.}{$\blacksquare$} d. s.} 
			& \rotatebox{90}{\textcolor{other flat}{$\blacksquare$} o. f.} 
			& \rotatebox{90}{\textcolor{sidewalk}{$\blacksquare$} sid.} 
			& \rotatebox{90}{\textcolor{terrain}{$\blacksquare$} ter.} 
			& \rotatebox{90}{\textcolor{manmade}{$\blacksquare$} man.} 
			& \rotatebox{90}{\textcolor{vegetation}{$\blacksquare$} veg.}
			& \makecell[c]{mIoU}
			\\
			% & mIoU\\
			\midrule
			MonoScene~\cite{cvpr22-monoscene} & \cmark & 1.8 & 7.2 & 4.3 & 4.9 & 9.4 & 5.7 & 4.0 & 3.0 & 5.9 & 4.5 & 7.2 & 14.9 & 6.3 & 7.9 & 7.4 & 1.0 & 7.7 & 6.06  \\
			
			TPVFormer~\cite{cvpr23-tpvformer} & \cmark & 7.2 & 38.9 & 13.7 & 40.8 & 45.9 & 17.2 & 20.0 & 18.9 & 14.3 & 26.7 & 34.2 & 55.7 & 35.5 & 37.6 & 30.7 & 19.4 & 16.8 & 27.83 \\
			
			OccFormer~\cite{arxiv23-occformer}  & \cmark & 5.9 & 30.3 & 12.3 & 34.4 & 39.2 & 14.4 & 16.5 & 17.2 & 9.3 & 13.9 & 26.4 & 51.0 & 31.0 & 34.7 & 22.7 & 6.8 & 7.0 & 21.93 \\

			CTF-Occ~\cite{arxiv23-occ3d}  & \cmark & 8.1 & 39.3 & 20.6 & 38.3 & 42.2 & 16.9 & 24.5 & 22.7 & 21.1 & 23.0 & 31.1 & 53.3 & 33.8 & 38.0 & 33.2 & 20.8 & 18.0 & 28.53 \\	

			BEVFormer~\cite{eccv22-bevformer}  & \cmark & 9.6 &  47.8 & 24.2  & 48.7  & 54.0 & 20.9 & 28.8 & 27.5 & 26.7 & 32.8 & 38.8 & 81.7 & 40.3 & 50.5  & 52.9 & 43.8  & 37.5 & 39.19 \\	

			BEVDet~\cite{arxiv21-bevdet}  & \cmark & 8.8 & 45.2 & 19.1 & 43.5 & 50.2 & 23.7 & 19.8 & 22.9 & 20.7 & 31.9 & 37.7 & 80.3 & 37.0 & 50.5 & 53.4 & 47.1 & 41.9 & 37.28 \\	

			\midrule

			SelfOcc-BEV~\cite{arxiv23-selfocc} & \xmark & 0.0 & 0.0 & 0.0 & 0.0 & 9.8 & 0.0 & 0.0 & 0.0 & 0.0 & 0.0 & 7.0 & 47.0 & 0.0 & 18.8 & 16.6 & 11.9 & 3.8 & 6.76 \\		

			SelfOcc-TPV~\cite{arxiv23-selfocc} & \xmark & 0.0 &0.0 & 0.0 & 0.0 & 10.0 & 0.0 & 0.0 & 0.0 & 0.0 & 0.0 & 7.11 & 53.0 & 0.0 & 23.6 & 25.2 & 12.0 & 4.6 & 7.97 \\	
			
			OccNeRF~\cite{arxiv23-occnerf}  & \xmark & 0.0 & 0.8 & 0.8 & 5.1 & 12.5 & 3.5 & 0.2 & 3.1 & 1.8 & 0.5 & 3.9 & 52.6 & 0.0 & 20.8 & 24.8 & 18.5 & 13.2 & 9.54 \\		

			\midrule
			VEON-B (Ours)  & \xmark & 0.5 & 4.8 & 2.7 & 14.7 & 10.9 & 11.0 & 3.8 & 4.7 & 4.0 & 5.3 & 9.6 & 46.5 & 0.7 &21.1 & 22.1 & 24.8 & 23.7 & 12.38 \\	
			
			VEON-L (Ours)  & \xmark & 0.9 & 10.4 & 6.2 & 17.7 & 12.7 & 8.5 & 7.6 & 6.5 & 5.5 & 8.2 & 11.8 & 54.5 & 0.4 & 25.5 & 30.2 & 25.4 & 25.4 & 15.14 \\		
			\bottomrule
	\end{tabular}}
    % \vspace{-15pt}
	\label{tab:base_main}
\end{table*}

In the sequel, we evaluate our VEON on the Occ3D-nuScenes validation set~\cite{arxiv23-occ3d}. We first report the 3D occupancy prediction results with either zero or partial manual semantic labels (\ie, either $C_{s} = \varnothing$ or $C_{s} \not= \varnothing$), and then prove the open-vocabulary capability of VEON both quantitatively and qualitatively. 

\myparagraph{Occupancy without semantic labels ($C_{s} = \varnothing$).} In Tab.~\ref{tab:base_main}, we investigate the performance of our VEON models trained without any manual semantic annotations. 
The first $6$ rows in Tab.~\ref{tab:base_main} list some supervised occupancy prediction models trained with full manual annotations. 
Performance of MonoScene~\cite{cvpr22-monoscene}, TPVFormer~\cite{cvpr23-tpvformer}, OccFormer~\cite{arxiv23-occformer}, CTF-Occ~\cite{arxiv23-occformer} is directly collected from \cite{arxiv23-occ3d}, while BEVFormer~\cite{eccv22-bevformer} and BEVDet~\cite{arxiv21-bevdet} are trained and evaluated on our own with the visible mask protocol~\cite{cvprw23-occupancychallenge}. On the other hand, rows $7$-$9$ in Tab.~\ref{tab:base_main} list three occupancy predictors trained without any manual annotations, including two variants of SelfOcc~\cite{arxiv23-selfocc} structured as  BEVFormer~\cite{eccv22-bevformer} and TPVFormer~\cite{cvpr23-tpvformer}, as well as the OccNeRF~\cite{arxiv23-occnerf} occupancy predictor. 
Finally, the last two rows demonstrate the performance of our VEON-B and VEON-L variants, which differ only in their CLIP backbones. Notice that our VEON utilizes the pseudo depth and the binary occupancy label from Occ3D~\cite{arxiv23-occ3d} for supervision, but these two can both be derived from the raw point cloud. 
From Tab.~\ref{tab:base_main}, we observe that our VEON-B and VEON-L respectively achieve a competitive performance of $12.38$ and $15.14$ mIoU. The VEON-L variant surpasses SelfOcc-BEV, SelfOcc-TPV, and OccNeRF respectively by $8.38$, $7.17$, and $5.60$ in mIoU. The performance boost of VEON comes from its capability of recognizing various objects, including some tail categories such as barrier, construction vehicles, bus and truck. For example, VEON-B and VEON-L obtain $4.8$ and $10.4$ IoU in barrier, while the corresponding performance for SelfOcc-BEV, SelfOcc-TPV and OccNeRF is only $0.0$, $0.0$ and $0.8$. Similar phenomena can be observed within other classes.

\myparagraph{Occupancy with partial semantic labels ($C_{s} \not= \varnothing$).} In the $C_{s} \not= \varnothing$ setting, we have $X$ seen classes with annotations and $Y$ unseen classes without annotations. 
In Tab.~\ref{tab:ov}, we select the VEON-L variant with two different $X/Y$ divisions ($X/Y=9/8$ and $X/Y=13/4$). The $X/Y=0/17$ variant is also listed as a baseline. The left $X$ and the right $Y$ classes are respectively seen and unseen classes~\cite{ral22-panopticnuscenes}. 
From Tab.~\ref{tab:ov}, we see that the mIoUs of VEON-L variants rise with $X$, which basically comes from the additional seen classes. Besides, the IoUs on unseen classes (\eg,  sidewalk, vegetation) are always competitive, contributing to the performance boost from another aspect. 

\definecolor{others}{RGB}{175, 0, 75}
\definecolor{barrier}{RGB}{255, 120, 50}
\definecolor{bicycle}{RGB}{255, 192, 203}
\definecolor{bus}{RGB}{255, 255, 0}
\definecolor{car}{RGB}{0, 255, 255}
\definecolor{const. veh.}{RGB}{255, 0, 255}
\definecolor{motorcycle}{RGB}{200, 180, 0}
\definecolor{pedestrian}{RGB}{255, 0, 0}
\definecolor{traffic cone}{RGB}{255, 240, 150}
\definecolor{trailer}{RGB}{135, 60, 0}
\definecolor{truck}{RGB}{160, 32, 240}
\definecolor{drive. suf.}{RGB}{139, 137, 137}
\definecolor{other flat}{RGB}{0, 150, 245}
\definecolor{sidewalk}{RGB}{75, 0, 75}
\definecolor{terrain}{RGB}{150, 240, 80}
\definecolor{manmade}{RGB}{230, 230, 250}
\definecolor{vegetation}{RGB}{0, 175, 0}

\begin{table*}[t]
	\setlength{\tabcolsep}{0.004\linewidth}
	\newcommand{\classfreq}[1]{{~\tiny(\semkitfreq{#1}\%)}}  %
	\centering
		\caption{Results on the Occ3D-nuScenes occupancy benchmark~\cite{cvpr20-nuscenes,arxiv23-occ3d} with $C_{s} \not= \varnothing$. }
  \label{tab:ov}
   % \vspace{-3pt}
	\resizebox{1\linewidth}{!}{
	\begin{tabular}{l|c c| c c c c c c c c c c c c c c c c c | c}
			
			\toprule
			Method
			& \rotatebox{90}{X: seen} & \rotatebox{90}{Y: uns.} 
			& \rotatebox{90}{\textcolor{others}{$\blacksquare$} oth.} 
			& \rotatebox{90}{\textcolor{barrier}{$\blacksquare$} bar.} 
			& \rotatebox{90}{\textcolor{bicycle}{$\blacksquare$} bic.}
			& \rotatebox{90}{\textcolor{bus}{$\blacksquare$} bus} 
			& \rotatebox{90}{\textcolor{car}{$\blacksquare$} car} 
			& \rotatebox{90}{\textcolor{const. veh.}{$\blacksquare$} c. v.} 
			& \rotatebox{90}{\textcolor{motorcycle}{$\blacksquare$} mot.} 
			& \rotatebox{90}{\textcolor{pedestrian}{$\blacksquare$} ped.} 
			& \rotatebox{90}{\textcolor{traffic cone}{$\blacksquare$} t. c.} 
			& \rotatebox{90}{\textcolor{trailer}{$\blacksquare$} tra.} 
			& \rotatebox{90}{\textcolor{truck}{$\blacksquare$} tru.} 
			& \rotatebox{90}{\textcolor{drive. suf.}{$\blacksquare$} d. s.} 
			& \rotatebox{90}{\textcolor{other flat}{$\blacksquare$} o. f.} 
			& \rotatebox{90}{\textcolor{sidewalk}{$\blacksquare$} sid.} 
			& \rotatebox{90}{\textcolor{terrain}{$\blacksquare$} ter.} 
			& \rotatebox{90}{\textcolor{manmade}{$\blacksquare$} man.} 
			& \rotatebox{90}{\textcolor{vegetation}{$\blacksquare$} veg.}
			& \makecell[c]{mIoU}
			\\
			
			\midrule
			
			VEON-L & 0 & 17 & 0.9 & 10.4 & 6.2 & 17.7 & 12.7 & 8.5 & 7.6 & 6.5 & 5.5 & 8.2 & 11.8 & 54.5 & 0.4 & 25.5 & 30.2 & 25.4 & 25.4 & 15.14 \\
			
			VEON-L & 9 & 8 & 0.9 & 14.3 & 4.4 & 26.6 & 15.0 & 7.5 & 7.4 & 5.6 & 5.0 & 8.3 & 9.2 & 48.7 & 0.1 & 24.9  & 30.6 & 24.8 & 24.5 & 15.16 \\
			
			VEON-L & 13 & 4 & 1.6 & 19.7 & 4.5 & 28.1 & 24.8 & 9.4 & 11.1 & 8.6 & 7.3 & 15.1 & 18.4 & 58.9 & 24.0 & 26.5 & 29.6 & 26.8 & 25.2 & 19.94 \\

			\bottomrule
	\end{tabular}}
% \vspace{-5pt}
\end{table*}
\begin{table}[t]
   \scriptsize
   	\setlength{\tabcolsep}{0.004\linewidth}
   \newcommand{\classfreq}[1]{{~\tiny(\semkitfreq{#1}\%)}}  %
    \hspace{8pt}
  	\begin{minipage}[b]{0.98\linewidth}
  		\centering
  		\makeatletter\def\@captype{table}
  		\setlength{\tabcolsep}{1mm}{
      		\caption{Results on the open-vocabulary language-driven retrieval benchmark~\cite{nips23-pop3d}.}
		\label{tab:retrieval}
		\begin{tabular}{c|cccccc}
			\toprule
			Method / mAP ($\%$)  & train (all) & train (vis) & val (all) & val (vis) & test (all) & test (vis) \\
			\midrule
			MaskCLIP+~\cite{eccv22-maskclip} & - & 13.5 & - & 18.7 & - &  12.0 \\
			POP-3D~\cite{nips23-pop3d} & 15.3 & 15.6 & 24.1 & 24.7 & 12.6 & 13.6 \\
			\midrule
			VEON-L (Ours) & \textbf{37.7} & \textbf{38.5} & \textbf{35.3} & \textbf{36.1} & \textbf{30.9} &  \textbf{31.3} \\
			\bottomrule
		\end{tabular}
     }
    \end{minipage}
% \vspace{-10pt}
\end{table}

\myparagraph{Open-vocabulary language-driven retrieval.} To quantitatively measure the open-vocabulary capability of our VEON, we evaluate our models on an open-vocabulary language-driven object retrieval benchmark proposed in \cite{nips23-pop3d}. Given an open-vocabulary language prompt, models need to retrieve relevant LiDAR points in the 3D space. 
The mean Average Precision (mAP) metric is used for evaluation, similar to the conventional retrieval problems. 
The benchmark provides annotations on $42$/$27$/$36$ scenes in the nuScenes training, validation, and testing set~\cite{cvpr20-nuscenes}. We strictly follow the open-sourced POP-3D codes to evaluate our VEON-L, with mAP on all points (mAP-all) and mAP on visible points (mAP-vis) as metrics. 
Note that our VEON-L variant is trained with zero semantic annotations ($X/Y=0/17$), and is never tuned on any retrieval labels. Tab.~\ref{tab:retrieval} gives the performance comparison between MaskCLIP+~\cite{eccv22-maskclip}, POP-3D~\cite{nips23-pop3d} and our VEON-L. Our VEON-L surpasses POP-3D by a significant margin, with $22.4\%$, $11.2\%$, $18.3\%$ mAP-all boost, and $22.9\%$,  $11.4\%$, $17.7\%$ mAP-vis boost on the training, validation and testing set, respectively. This suggests that the 3D representation output from VEON aligns well with language embeddings of CLIP, with powerful capability of handling open-vocabulary tasks.

\begin{figure*}[t]
	\centering
 \includegraphics[width=0.87\linewidth,keepaspectratio]{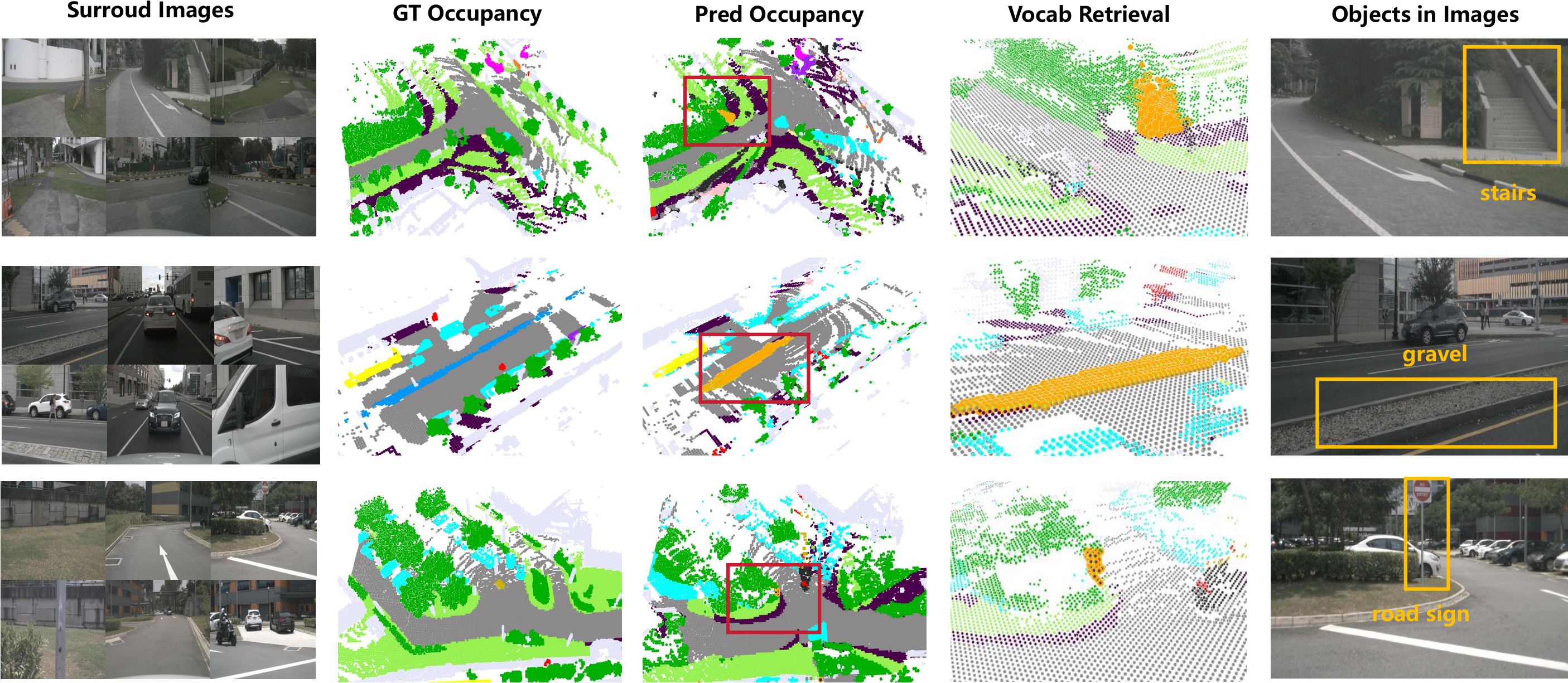}
	\caption{Visualization of occupancy prediction (VEON-L) on the Occ3D-nuScenes occupancy benchmark~\cite{cvpr20-nuscenes,arxiv23-occ3d} (validation set). We visualize the surrounding images (column $1$), ground truth and predicted occupancy (column $2$-$3$), and the retrieval results of certain open-vocabulary classes (column $4$-$5$). Our VEON-L demonstrates the capability of recognizing unseen objects (in orange), such as stairs, gravel, and road signs.}
	\label{fig:vis}
\end{figure*}

\myparagraph{Visualization.} 
In Fig.~\ref{fig:vis}, we qualitatively show the open-vocabulary capability of our VEON. Here, we choose the VEON-L variant with $X/Y=0/17$, meaning that all the visualized results are obtained without any manual semantic labels. 
We collect three scenes from the Occ3D-nuScenes~\cite{cvpr20-nuscenes,arxiv23-occ3d} validation set, and visualize one on each row.
As in Fig.~\ref{fig:vis}, column $1$ shows the surrounding images, and columns $2$-$3$ compare the occupancy results of ground truth and our VEON predicted ones. We see that our VEON shows promising results, keeping a great alignment with the ground truth. Columns $4$-$5$ are illustrations of open-vocabulary retrieval tests. Specifically, we utilize language embeddings of unseen classes in the vocabulary to find out which voxels in 3D space belong to the class. Notice that the classes for retrieval tests are fine-grained ``subclasses'' instead of ``superclasses'' defined by nuScenes (see the supplementary material for details). From Fig.~\ref{fig:vis}, we observe that our VEON succeeds in recognizing open-vocabulary classes such as stairs, gravel, and road sign.  This proves the efficacy of our model in recognizing open-world objects on the road. 

\subsection{Ablation Study}
\label{sec:ablation}

\myparagraph{Depth estimation.} VEON requires a depth estimation module $\phi_{dp}$, which is trained with low-rank adaptation (LoRA) for robust domain transfer. We conduct a thorough study on it in Tab.~\ref{tab:depth}. We follow the $C_{s} = \varnothing$ setting, with ViT-B and ViT-L as the CLIP backbone. We only report the IoUs of five representative classes (barrier, bicycle, pedestrian, truck, and vegetable) and the overall mIoU of all $17$ classes. 
From Tab.~\ref{tab:depth}, we discover that for the two variants of VEON, the mIoU rises moderately by $0.81$ and $0.57$. We conclude that preciser depth estimation module $\phi_{dp}$ leads to preciser 3D occupancy prediction results.

\begin{table}[t]
   \scriptsize
   \scalebox{0.95}{
  	\begin{minipage}[b]{0.49\linewidth}
  		\centering
  		\makeatletter\def\@captype{table}
  		\setlength{\tabcolsep}{1mm}{
      		\caption{Ablation study on training the depth module $\phi_{dp}$ via low rank adaptation. }
      	% \vspace{-5pt}
  		\label{tab:depth}
            \begin{tabular}{cc|ccccc|c}
				\toprule
				Variant & LoRA &  \rotatebox{90}{\textcolor{barrier}{$\blacksquare$} bar.}  & 
				\rotatebox{90}{\textcolor{bicycle}{$\blacksquare$} bic.} & 
				\rotatebox{90}{\textcolor{pedestrian}{$\blacksquare$} ped.}  & 
				\rotatebox{90}{\textcolor{truck}{$\blacksquare$} tru.} & 
				\rotatebox{90}{\textcolor{vegetation}{$\blacksquare$} veg.} & \makecell[c]{mIoU} \\
				\midrule
				VEON-B & \xmark & 4.7 & 3.6 & 5.1 & 7.3 & 23.2  & 11.57  \\
				VEON-B & \cmark & 4.8 & 2.7 & 4.7 & 9.6 & 23.7 & 12.38  \\
				\midrule
				VEON-L & \xmark & 10.2 & 4.9 & 3.0 & 12.4 & 22.8  & 14.57  \\
				VEON-L & \cmark & 10.4 & 6.2 & 6.5 & 11.8 & 25.4 & 15.14  \\
				\bottomrule
			\end{tabular}
    	}
  		% \vspace{-10pt}
    \end{minipage}
    \quad
    \hspace{8pt}
  	\begin{minipage}[b]{0.49\linewidth}
  		\centering
  		\makeatletter\def\@captype{table}
  		\setlength{\tabcolsep}{1mm}{
      		\caption{Ablation study on the class reweighting strategy for the tail class trap. }
      		% \vspace{-5pt}
  		\label{tab:loss}
            \begin{tabular}{cc|ccccc|c}
				\toprule
				Variant & RW &  \rotatebox{90}{\textcolor{barrier}{$\blacksquare$} bar.}  & 
				\rotatebox{90}{\textcolor{bicycle}{$\blacksquare$} bic.} & 
				\rotatebox{90}{\textcolor{pedestrian}{$\blacksquare$} ped.} & 
				\rotatebox{90}{\textcolor{truck}{$\blacksquare$} tru.} & 
				\rotatebox{90}{\textcolor{vegetation}{$\blacksquare$} veg.} & \makecell[c]{mIoU} \\
				\midrule
				VEON-B & \xmark & 4.4 & 0.0 & 0.0 & 8.3 & 25.5  & 10.39  \\
				VEON-B & \cmark & 4.8 & 2.7 & 4.7 & 9.6 & 23.7 & 12.38    \\
				\midrule
				VEON-L & \xmark & 8.9 & 4.9 & 3.5 & 13.6 & 25.5  & 14.18 \\
				VEON-L & \cmark & 10.4 & 6.2 & 6.5 & 11.8 & 25.4 & 15.14 \\
				\bottomrule
			\end{tabular}
}
  		% \vspace{-8pt}
    \end{minipage}}
    % \vspace{-18pt}
\end{table}

\myparagraph{Class reweighting.} We propose the class reweighting strategy (see Eq.~\ref{eq:reweight}) to escape from the tail class trap in open-vocabulary occupancy prediction. Experiments are conducted in Tab.~\ref{tab:loss} on both variants of VEON. We observe that adding the class reweighting strategy brings an increase of $1.99$ and $0.96$ in terms of mIoU. The reason can be found in the class-wise IoUs. Take the VEON-B variant as an example. After integrating the strategy, the IoU of ``bicycle'' rises from $0.0$ to $2.7$, and the IoU of ``pedestrian'' rises from $0.0$ to $4.7$. This indicates that the class reweighting strategy enables the network to recognize tail classes. 

\myparagraph{High-resolution side adaptor (HSA).} We test the indispensability of the HSA module in Tab.~\ref{tab:hsa}. Given our baseline as VEON-B, we first try a trivial solution of directly lifting the CLIP feature as in row $2$. The $0.70$ mIoU decrease proves the efficacy of adapting the CLIP foundation model. 
Then, we remove the attention bias matrix $\mathbf{A}$ and the supplementary matrix $\mathbf{S}$ respectively. As on rows $3$-$4$ in Tab.~\ref{tab:hsa}, we suffer from  $0.63$ and $0.77$ mIoU decrease. This indicates that our HSA succeeds in refining the CLIP features and provides a high-resolution supplement for semantic extraction. In row $5$, we try another solution of linearly predicting and adding offsets to CLIP visual tokens. The $0.32$ decrease infers that our attention bias solution is more feasible. 

\begin{table}[t]
   \scriptsize
   \scalebox{0.98}{
  	\begin{minipage}[b]{0.53\linewidth}
  		\centering
  		\makeatletter\def\@captype{table}
  		\setlength{\tabcolsep}{1mm}{
      		\caption{Ablation study on the High-resolution Side Adaptor (HSA) module. We remove ($-$) / add ($+$) features from/to the VEON-B to check component efficacy. }
  		% \vspace{-5pt}
  		\label{tab:hsa}
            \begin{tabular}{c|ccccc|c}
            \toprule
            Variant  &  \rotatebox{90}{\textcolor{barrier}{$\blacksquare$} bar.}  & 
            \rotatebox{90}{\textcolor{bicycle}{$\blacksquare$} bic.} & 
            \rotatebox{90}{\textcolor{pedestrian}{$\blacksquare$} ped.} & 
            \rotatebox{90}{\textcolor{truck}{$\blacksquare$} tru.} & 
            \rotatebox{90}{\textcolor{vegetation}{$\blacksquare$} veg.} & \makecell[c]{mIoU} \\
            \midrule
            VEON-B & 4.8 & 2.7 & 4.7 & 9.6 & 23.7 & 12.38  \\
            \midrule
            -- Whole HSA & 4.1 & 4.0 & 4.0 & 9.0 & 24.7 & 11.68 \\
            -- Attention bias $\mathbf{A}$ & 4.1 & 3.0 & 4.7 & 9.3 & 23.7 & 11.75 \\
            -- Supplementary $\mathbf{S}$ & 4.0 & 3.8 & 4.0 & 7.7 & 24.4 & 11.61 \\
            + Token Offsets & 4.2 & 3.7 & 5.1 & 9.3 & 24.4 & 12.06 \\
            \bottomrule
        \end{tabular}
    	}

%  		\vspace{-8pt}
    \end{minipage}
    \hspace{8pt}
  	\begin{minipage}[b]{0.46\linewidth}
  		\centering
  		\makeatletter\def\@captype{table}
  		\setlength{\tabcolsep}{1mm}{
      		\caption{Statistical results on the parameters (M),  trainable parameters (M) and the trainable fraction (\%) within each component in VEON-L. }
  		% \vspace{-5pt}
  		\label{tab:params}
	\begin{tabular}{cccc}
		\toprule
		Models & Param & Param-Tr. & Frac.\\
		\midrule
		MiDaS $\phi_{rel}$ & 328.7 & 0.9 & 0.3\% \\
		D-Adaptor $\phi_{r2m}$ & 16.8 & 16.8 & 100.0\% \\
		D-Model $\phi_{dp}$ & 345.5 & 17.7 & 5.1\%\\
		\midrule
		CLIP ViT-L & 304.3 & 0 & 0\% \\
		HSA & 13.5 & 13.5 & 100.0\% \\
		3D Layers & 14.8 & 14.8 & 100.0\% \\
		\midrule
		VEON-L & 678.1 & 46.0 &  6.8\% \\
		\bottomrule
	\end{tabular}
     }
    \end{minipage}
    }
    % \vspace{-20pt}
\end{table}

\myparagraph{Parameter statistics.} From Tab.~\ref{tab:params}, we do statistics for the parameters of each component within our VEON-L model. While our model has a tremendous parameter number of $678.1$M due to the integration of two foundation models, the trainable parts within our VEON only occupy a small fraction of $6.8\%$, with $17.7$M in the depth module (stage $1$) and $28.3$M in the occupancy predictor (stage $2$). This affirms that our VEON remains lightweight.

\section{Concluding Remarks}
\label{sec:conclusion}

In this paper, we design a \ownmethod framework for \textbf{V}ocabulary-\textbf{E}nhanced \textbf{O}ccupancy predictio\textbf{N}. 
We adopt a decoupled structure for 3D occupancy prediction, which assembles a depth foundation model MiDaS and a semantic foundation model CLIP. As directly integrating these two models meets challenges, we adapt MiDaS with a relative-metric-bin adaptor and low-rank adaptation (LoRA) for domain transfer, and equip CLIP with a high-resolution side adaptor (HSA) for enhanced feature extraction. We also design a class reweighting loss to escape from the tail class trap. Our VEON method shows competitive performance on the Occ3D-nuScenes dataset and strong capability of recognizing unseen and fine-grained classes. We hope our work could herald a rethinking of the construction pipeline of open-vocabulary 3D occupancy prediction models. 

\noindent\textbf{Acknowledgements.} This work was supported by NSFC (62322113, 62376156), Shanghai Municipal Science and Technology Major Project (2021SHZDZX0102), and the Fundamental Research Funds for the Central Universities.

{
	\bibliographystyle{splncs04}
	\bibliography{ms}
}

\appendix
\newpage

\vspace{5pt}
\begin{center}
	\textbf{\Large Supplementary Material}
\end{center}
\vspace{7pt}

\renewcommand{\thefigure}{\Alph{figure}}
\renewcommand{\thetable}{\Alph{table}}
\renewcommand{\thesection}{\Alph{section}}
\renewcommand{\theequation}{\Alph{equation}}
\setcounter{section}{0}
\setcounter{table}{0}
\setcounter{equation}{0}
\setcounter{figure}{0}

In the supplementary material, we first present some details of our VEON framework, including class embedding generation, subclass division, depth loss, feature alignment, and attention bias. 
Then, we provide more quantitative results and visualization on the nuScenes~\cite{cvpr20-nuscenes} dataset to demonstrate the open-vocabulary capability of our VEON. Finally, we discuss the potential negative societal impact and limitations of our work. 

\section{Framework Details}
\label{sec:details}

\subsection{Class Embedding Generation} 

In our VEON framework, we align the voxel-wise semantic-aware occupancy map $\mathbf{O^{sa}}$ with the CLIP~\cite{icml21-clip} language embeddings of specific classes, formulated as Eq.~\ref{eq:reweight} in the manuscript. 
To generate class embeddings suitable for open-vocabulary recognition, we combine multiple natural language templates to jointly describe each single class. We then average the corresponding embeddings output from the CLIP language encoder to obtain the required embedding for each class~\cite{cvpr23-ovsegmaskadclip,eccv22-simseg}. In practice, $14$ templates are collected following SAN~\cite{cvpr23-san}. An example is ``This is a photo of a \{\}'', where \{\} represents the class name text. Tab.~\ref{tab:prompt} shows the detailed list of the prompt templates. 

\begin{table}[h]
	\small
	\centering
	\begin{tabular}{l}
		\toprule
		``a photo of a \{\}.", \\
		``This is a photo of a \{\}",\\
		``There is a \{\} in the scene",\\
		``There is the \{\} in the scene",\\
		``a photo of a \{\} in the scene",\\
		``a photo of a small \{\}.",\\
		``a photo of a medium \{\}.",\\
		``a photo of a large \{\}.",\\
		``This is a photo of a small \{\}.",\\
		``This is a photo of a medium \{\}.",\\
		``This is a photo of a large \{\}.",\\
		``There is a small \{\} in the scene.",\\
		``There is a medium \{\} in the scene.",\\
		``There is a large \{\} in the scene.",\\
		\bottomrule
	\end{tabular}
        \vspace{5pt}
	\caption{List of prompt templates used in VEON. We keep the same templates as those utilized in SAN~\cite{cvpr23-san}.}
	\label{tab:prompt}
\end{table}

\subsection{Subclass Division}  

\begin{table}[ht]
	\small
	\centering
	\begin{tabularx}{1.0\textwidth}{l|X}
		\toprule
		Superclass & List of subclasses \\
		\midrule
		others & debris, animal, personal mobility, skateboard, segway, scooter, stroller, wheelchair, trash bag, trash can, wheelbarrow, bicycle rack, ambulance, police vehicle. \\
		barrier & traffic barrier. \\
		bicycle & bicycle. \\
		bus & bus. \\
		car & car, sedan, hatch-back, wagon, van, SUV, jeep. \\ 
		const. veh. & construction vehicle. \\
		motorcycle & motorcycle. \\
		pedestrian & pedestrian, construction worker, police officer. \\ 
		traffic cone & traffic cone. \\
		trailer & trailer. \\
		truck & truck. \\
		driv. surf. & road. \\ 
		other flat & traffic island, traffic delimiter, rail track, lake, river.\\
		sidewalk & sidewalk, pedestrian walkway, bike path. \\
		terrain & grass, rolling hill, soil, sand, gravel. \\
		manmade & building, wall, guard rail, fence, drainage, hydrant, banner, street sign, traffic light, parking meter, stairs. \\
		vegetation & vegetation, plants, bushes, tree. \\
		\bottomrule
	\end{tabularx}
	\vspace{5pt}
	\caption{The subclass list used in VEON. The superclasses are kept the same as the predefined classes in nuScenes~\cite{cvpr20-nuscenes,ral22-panopticnuscenes}, and the subclasses are summarized from the official class description from the nuScenes LiDAR segmentation~\cite{ral22-panopticnuscenes} benchmark.}
	\label{tab:subclass}
\end{table}

In VEON, we need to define an overall class set $C_{all}$ for open-vocabulary recognition. 
The selection of $C_{all}$ seems to be trivial at first glance, as Occ3D-nuScenes~\cite{cvpr20-nuscenes,arxiv23-occ3d} natively classifies all voxels into $17$ non-free classes~\cite{ral22-panopticnuscenes} and $1$ free class. 
However, we find such coarse-grained class division unsuitable for open-vocabulary tasks. 
For example, the first non-free class in Occ3D-nuScenes is termed as ``others'', obviously a meaningless class description. Voxels labeled as ``others'' may be occupied by various subclasses of objects, including animal, trash can, skateboard, personal mobility, and ego vehicle, etc. Therefore, using the coarse-grained class terms provided by Occ3D-nuScenes is improper.

To better suit the class embeddings to the open-vocabulary task, we adopt a \emph{subclass division strategy} that divides the original superclasses collected from Occ3D-nuScenes into separate subclasses. This enlarges the overall (non-free) class set $C_{all}$ from the original $17$ superclasses to $\sim 60$ subclasses. The detailed list of subclasses, summarized from the official nuScenes description of these coarse superclasses, is shown in Tab.~\ref{tab:subclass}. 

With the subclass division strategy, we achieve a fine-grained understanding of the surrounding 3D space during inference. For instance, tree, bushes and other plants could be distinguished into different subclasses, despite that they all belong to the superclass ``vegetation''.
Note that for quantitative evaluation on the Occ3D-nuScenes benchmark, we project the subclasses back to the superclasses according to Tab.~\ref{tab:subclass}, and calculate the class-wise IoUs and overall mIoU metrics.

\subsection{Depth Loss}

In the first stage of VEON, we supervise the metric depth map $\mathbf{D}$ with a pixel-wise scale-invariant depth loss $L_{pix}$. Suppose $d_{i}$ is the $i$-th pixel of $\mathbf{D}$, and $\hat{d}_{i}$ is the $i$-th pixel of the corresponding ground truth $\mathbf{\hat{D}}$. Here $\mathbf{\hat{D}}$ is obtained by projecting the point cloud $\textbf{P}$ onto the camera plane. Then, we strictly follow \cite{nips14-depth,arxiv23-zoedepth,cvpr21-adabins} to calculate the pixel-wise scale-invariant depth loss $L_{pix}$ as:
\begin{equation}\label{eq:pixel_loss}
	L_{pix} = \sqrt{\frac{1}{N_{pix}} \sum_{i} g_{i}^{2} -  \frac{\alpha}{N_{pix}^{2}} \left(\sum_{i} g_{i} \right)^{2}},
\end{equation} 
where $N_{pix}$ is the total number of pixels on $\mathbf{D}$, $\alpha$ is a constant, and $g_{i}$ is the log-difference between each depth $d_{i}$ and its corresponding ground truth $\hat{d}_{i}$ on $\mathbf{\hat{D}}$, namely $g_{i} = \log{d_{i}} - \log{\hat{d}_{i}}$. As is explained in the manuscript, $L_{pix}$ ensures the shape and smoothness of the output metric depth map $\mathbf{D}$. This design helps retain knowledge from the depth foundation model MiDaS~\cite{tpami20-midas}, and is also beneficial to the subsequent bin depth transformation. As an implementation detail, $L_{pix}$ is calculated on the $8\times$-downsampled depth maps compared with the input surrounding images. Also, for those pixels without pseudo depth projected from the point cloud $\textbf{P}$, they will never be involved in loss calculation.

\subsection{Feature Alignment} 

In VEON, we align the semantic-aware occupancy map $\mathbf{O^{sa}}$ with existing 2D pixel-wise CLIP-aligned embeddings, as Eq.~\ref{eq:reweight} in the manuscript. 
We design to utilize an off-the-shelf 2D open-vocabulary segmentor SAN~\cite{cvpr23-san} to generate the 2D pixel-wise CLIP-aligned embeddings. Then, $\mathbf{O^{sa}}$ is supervised via 3D-to-2D projection and feature alignment. We will dive into detail in the sequel. 

First, we introduce how to generate the 2D CLIP-aligned embeddings with SAN~\cite{cvpr23-san}. SAN is an open-vocabulary 2D segmentor composed of a CLIP image encoder and a side adaptor network. It utilizes a query-based methodology to generate (1) class-agnostic object mask proposals and (2) proposal-wise embeddings by manipulating the CLIP attention layers. The final output of SAN is a pixel-wise classification map for the input 2D surrounding images. On each pixel, $\left| C_{all} \right|$ probabilities are given, indicating the likelihood that the pixel belongs to each particular class. 
For the detailed architecture of SAN, we refer readers to \cite{cvpr23-san}. 

Second, we present details of the feature alignment process. For each voxel in the 3D space, we first project the center of the voxel onto the surrounding images based on the intrinsic and extrinsic camera parameters. The following procedure shifts according to the availability of semantic label on the voxel. 
If there exists no superclass label on the voxel, we select the subclass in $C_{all}$ with the highest classification probability on the projected pixel, and pick the corresponding CLIP language embedding as the (pseudo) ground truth for that voxel. If there exists a superclass label on the voxel (typically when $C_{s} \not= \varnothing$), we select the subclass restricted by the superclass annotation, and other procedures are kept the same. For example, consider a 3D voxel labeled as the superclass ``vegetation''. We refer to the projected 2D pixel on surrounding images and fetch the output of SAN on that pixel. In this case, only $4$ subclasses, including ``vegetation'', ``plants'', ``bushes'' and ``tree'' will be regarded as candidate subclasses (see Tab.~\ref{tab:subclass}), and the single subclass with the highest classification probability will be selected as pseudo ground truth class for supervising $\mathbf{O^{sa}}$. The class embedding to align is then fetched from the CLIP language encoder.

\subsection{Attention Bias} 

We design a High-resolution Side Adaptor (HSA) to make the pretrained CLIP better suited to the open-vocabulary occupancy prediction task. The key idea is to maintain a side adaptor that absorbs early layers of visual tokens from CLIP and then outputs an attention bias matrix $\mathbf{A}$ to manipulate the attention layers in the later layers of CLIP. The HSA module has a higher resolution than the CLIP backbone, contributing to fine-grained scene understanding by providing high-resolution supplementary information. 

Here, we focus on how the attention bias matrix $\mathbf{A}$ affects the forward pipeline of CLIP transformer layers. The CLIP backbone follows the ViT~\cite{arxiv20-vit} architecture. Images are sliced into patches of $16 \times 16$, encoded into initial visual tokens $\mathbf{X}_{0}^{[v]}$, and concatenated with an initial global [cls] token $\mathbf{X}_{0}^{[cls]}$. The tokens $\mathbf{X}_{0} = [\mathbf{X}_{0}^{[v]}, \mathbf{X}_{0}^{[cls]}]$ go through multiple transformer layers (12/24 layers for ViT-B/ViT-L), where the operation $\left[\cdot, \cdot \right]$ means token concatenation. Each transformer layer comprises multi-head attention, feed-forward network, and layer normalization~\cite{arxiv20-vit}. Our attention bias matrix $\mathbf{A}$ operates solely in the multi-head attention. In the manuscript, we simplify the process as follows (copied from Eq.~\ref{eq:attention_bias} in the manuscript): 
\begin{equation}\label{eq:attention_bias_supp}
\mathbf{X}_{i+1} = \operatorname{softmax}(\mathbf{Q}_{i}\mathbf{K}_{i}^{T} + \mathbf{A}_{i} \mathbf{A}_{i}^{T}) \mathbf{V}_{i}.
\end{equation}
Here $\mathbf{X}_{i}$ represents the visual tokens in the $i^{th}$ layer, and  $\mathbf{Q}_{i}$, $\mathbf{K}_{i}$ and $\mathbf{V}_{i}$ are the linear transformations of $\mathbf{X}_{i}$. The attention bias $\mathbf{A}_{i} \mathbf{A}_{i}^{T}$ for the $i^{th}$ layer is added to $\mathbf{Q}_{i}\mathbf{K}_{i}^{T}$ for directing the transformer to pay more attention on the spatial information. 

\begin{figure*}[t]
	\centering
	\includegraphics[width=0.32\linewidth,keepaspectratio]{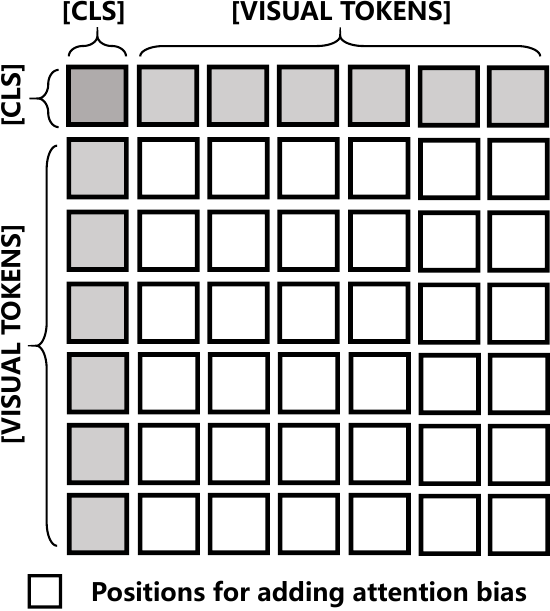}
	\caption{Positions for adding attention bias (blank squares).}
	\label{fig:add_bias}
\end{figure*}

In fact, we ignore three details in the above formulation. \textit{First}, in Eq.~\ref{eq:attention_bias_supp}, we omit the feed-forward network and layer normalization in each transformer layer. In other words, the output in Eq.~\ref{eq:attention_bias_supp} should additionally pass through the feed-forward network and layer normalization to become the input tokens $\mathbf{X}_{i+1}$ of the next transformer layer. \textit{Second}, the global [cls] token $\mathbf{X}_{i}^{[cls]}$ is ignored in Eq.~\ref{eq:attention_bias_supp}. As is shown in Fig.~\ref{fig:add_bias}, each attention operation involves the feature interaction between $\mathbf{X}_{i}^{[v]}$ and $\mathbf{X}_{i}^{[cls]}$. Our attention bias $\mathbf{A}_{i}$ for layer $i$ is added only to the attention parts of visual tokens, \ie, the blank positions in Fig.~\ref{fig:add_bias}. Also, the scale constant $\frac{1}{\sqrt{d}}$ is also omitted in Eq.~\ref{eq:attention_bias_supp} ($d$ is the dimension). \textit{Third}, multiple attention heads are calculated separately in each transformer layer. In our VEON, the attention biases are also separate for each head. This means that the HSA head needs to output the attention bias for all the attention heads in all the later layers of CLIP. For example, in the ViT-L CLIP, the attention bias matrix $\mathbf{A}$ has a size of $(\frac{H}{16} \times \frac{W}{16}) \times 6 \times 8 \times 32$. Here $H$ and $W$ are the height and width of the input image, $6$ is the number of layers being manipulated by $\mathbf{A}$, and $8$ is the number of heads in each multi-head attention. Then, the inner production within $\mathbf{A}\mathbf{A}^{T}$ in Eq.~\ref{eq:attention_bias_supp} is performed on the last dimension of $\mathbf{A}$, with the head dimension as $32$. In other words, $\mathbf{A} \mathbf{A}^{T}$ has the size of $(\frac{H}{16} \times \frac{W}{16}) \times  (\frac{H}{16} \times \frac{W}{16}) \times 6 \times 8$, indicating the layer-wise and head-wise attention biases in the transformer layers.

\section{More Experimental Results}

\subsection{Occupancy Prediction with $C_{s} \not= \varnothing$}

In Tab.~\ref{tab:ov} in the manuscript, we investigate the occupancy prediction performance of our VEON-L in the $C_{s} \not= \varnothing$ setting. Here we repeat the experiment on another variant, namely VEON-B, in Tab.~\ref{tab:ov_base}. Remember that with $C_{s} \not= \varnothing$, we have $X$ seen classes with semantic annotations and $Y$ unseen classes without semantic annotations. 
Similar to Tab.~\ref{tab:ov}, we pick two different $X/Y$ divisions ($X/Y=9/8$ and $X/Y=13/4$), and the $X/Y=0/17$ variant is also provided for comparison. Note that the left $X$ and the right $Y$ classes in Tab.~\ref{tab:ov_base} are seen and unseen classes~\cite{ral22-panopticnuscenes}, respectively. 
In other words, in the $X/Y=9/8$ case, classes from ``others'' to ``traffic cone'' are seen classes, while the classes from ``trailer'' to ``vegetation'' are unseen classes.

From Tab.~\ref{tab:ov_base}, we observe three phenomena. \textit{First}, similar to the results of VEON-L, the VEON-B variant also benefits from the increase in seen classes $X$. When $X$ rises from $0 \rightarrow 9 \rightarrow 13$, the mIoU also increases from $12.38 \rightarrow 13.26 \rightarrow 17.38$. This overall mIoU increase primarily comes from the additional seen classes, such as the $14.7 \rightarrow 23.8 \rightarrow 26.8$ IoU increase in the class ``bus'', while the performance on unseen classes remains stable. \textit{Second}, comparing Tab.~\ref{tab:ov} with Tab.~\ref{tab:ov_base}, we discover that with all three types of $X/Y$ settings, the VEON-L variants surpass the VEON-B variants respectively by $2.76$, $1.90$, and $2.56$ mIoU. This affirms that 2D data prior originating from large-scale vision language pretraining is critical for 3D open-vocabulary tasks such as occupancy prediction. \textit{Third}, the VEON variants do not perform well on certain classes when they are not explicitly annotated, \eg, ``other flats''. This can attributed to the failure of the open-vocabulary segmentor SAN~\cite{cvpr23-san} in recognizing superclass ``other flats'', which includes stuff subclasses such as traffic island, traffic delimiter, river, etc.

\definecolor{others}{RGB}{175, 0, 75}
\definecolor{barrier}{RGB}{255, 120, 50}
\definecolor{bicycle}{RGB}{255, 192, 203}
\definecolor{bus}{RGB}{255, 255, 0}
\definecolor{car}{RGB}{0, 255, 255}
\definecolor{const. veh.}{RGB}{255, 0, 255}
\definecolor{motorcycle}{RGB}{200, 180, 0}
\definecolor{pedestrian}{RGB}{255, 0, 0}
\definecolor{traffic cone}{RGB}{255, 240, 150}
\definecolor{trailer}{RGB}{135, 60, 0}
\definecolor{truck}{RGB}{160, 32, 240}
\definecolor{drive. suf.}{RGB}{139, 137, 137}
\definecolor{other flat}{RGB}{0, 150, 245}
\definecolor{sidewalk}{RGB}{75, 0, 75}
\definecolor{terrain}{RGB}{150, 240, 80}
\definecolor{manmade}{RGB}{230, 230, 250}
\definecolor{vegetation}{RGB}{0, 175, 0}

\begin{table*}[t]
	\setlength{\tabcolsep}{0.004\linewidth}
	\newcommand{\classfreq}[1]{{~\tiny(\semkitfreq{#1}\%)}}  %
	\centering
		\caption{Performance of our VEON-B on the Occ3D-nuScenes occupancy benchmark~\cite{cvpr20-nuscenes,arxiv23-occ3d} in the $C_{s} \not= \varnothing$ setting. }
      \label{tab:ov_base}
	\resizebox{1\linewidth}{!}{
	\begin{tabular}{l|c c| c c c c c c c c c c c c c c c c c | c}
			
			\toprule
			Method
			& \rotatebox{90}{X: seen} & \rotatebox{90}{Y: uns.} 
			& \rotatebox{90}{\textcolor{others}{$\blacksquare$} oth.} 
			& \rotatebox{90}{\textcolor{barrier}{$\blacksquare$} bar.} 
			& \rotatebox{90}{\textcolor{bicycle}{$\blacksquare$} bic.}
			& \rotatebox{90}{\textcolor{bus}{$\blacksquare$} bus} 
			& \rotatebox{90}{\textcolor{car}{$\blacksquare$} car} 
			& \rotatebox{90}{\textcolor{const. veh.}{$\blacksquare$} c. v.} 
			& \rotatebox{90}{\textcolor{motorcycle}{$\blacksquare$} mot.} 
			& \rotatebox{90}{\textcolor{pedestrian}{$\blacksquare$} ped.} 
			& \rotatebox{90}{\textcolor{traffic cone}{$\blacksquare$} t. c.} 
			& \rotatebox{90}{\textcolor{trailer}{$\blacksquare$} tra.} 
			& \rotatebox{90}{\textcolor{truck}{$\blacksquare$} tru.} 
			& \rotatebox{90}{\textcolor{drive. suf.}{$\blacksquare$} d. s.} 
			& \rotatebox{90}{\textcolor{other flat}{$\blacksquare$} o. f.} 
			& \rotatebox{90}{\textcolor{sidewalk}{$\blacksquare$} sid.} 
			& \rotatebox{90}{\textcolor{terrain}{$\blacksquare$} ter.} 
			& \rotatebox{90}{\textcolor{manmade}{$\blacksquare$} man.} 
			& \rotatebox{90}{\textcolor{vegetation}{$\blacksquare$} veg.}
			& \makecell[c]{mIoU}
			\\
			% & mIoU\\
			\midrule

			VEON-B & 0 & 17 & 0.5 & 4.8 & 2.7 & 14.7 & 10.9 & 11.0 & 3.8 & 4.7 & 4.0 & 5.3 & 9.6 & 46.5 & 0.7 &21.1 & 22.1 & 24.8 & 23.7 & 12.38 \\
			
			VEON-B & 9 & 8 & 1.0 & 9.5 & 3.5 & 23.8 & 16.3 & 9.3 & 5.47 & 3.5 & 4.7 & 5.1 & 6.7 & 45.0 & 0.6 & 21.1 & 21.8 & 24.0 & 24.2 & 13.26 \\
			
			VEON-B & 13 & 4 & 0.9 & 9.5 & 4.8 & 26.8 & 25.7 & 10.4 & 7.9 & 5.2 & 9.4 & 10.1 & 16.4 & 62.0 & 14.7 & 23.4 & 19.3 & 24.6 & 24.5 & 17.38 \\

			\bottomrule
	\end{tabular}}
	
%	\vspace{-4mm}
\end{table*}

\subsection{More Visualization}
\label{sec:vis_supp}

\begin{figure*}[t]
	\centering
	\includegraphics[width=1.0\linewidth,keepaspectratio]{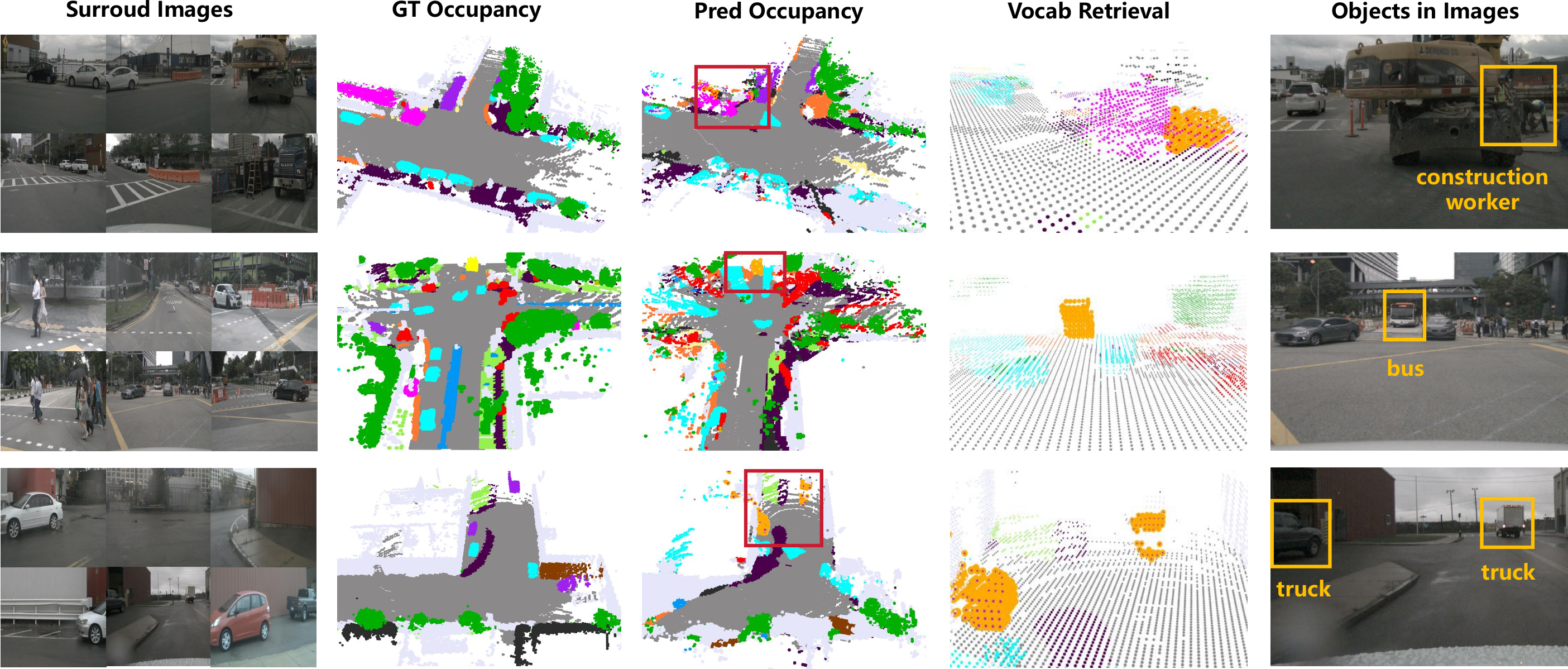}
	\caption{More visualization of occupancy prediction (VEON-L) on the Occ3D-nuScenes occupancy benchmark~\cite{cvpr20-nuscenes,arxiv23-occ3d} (validation set). We visualize the surrounding images (column $1$), ground truth and predicted occupancy (column $2$-$3$), and the open-vocabulary retrieval results of certain classes (column $4$-$5$). 	
		We see that our VEON-L not only shows competitive occupancy prediction results but also succeeds in recognizing unseen objects (colored in orange), such as construction worker, bus, truck, etc. Remember that the above results are obtained without any semantic labels. }
	\label{fig:vis_supp}
\end{figure*}

In Fig.~\ref{fig:vis_supp}, we qualitatively show the open-vocabulary capability of our VEON, as a supplement to Fig.~\ref{fig:vis} in the manuscript. All settings are kept the same as Fig.~\ref{fig:vis}, with VEON-L as our model and the Occ3D-nuScenes~\cite{cvpr20-nuscenes,arxiv23-occ3d} dataset as the benchmark. Remember that the selected VEON-L is trained without any semantic labels. 
In Fig.~\ref{fig:vis_supp}, column $1$ shows the surrounding images, and columns $2$-$3$ compare the ground truth occupancy and our VEON predicted ones. Columns $4$-$5$ visualize the open-vocabulary voxel retrieval results. 
Specifically, we utilize language embedding of any unseen subclass in $C_{all}$ to search for which voxels in 3D space belong to that subclass. 
Each occupancy snapshot in column $4$ is an enlarged view of the local occupancy in the red box in column $3$, and the camera image in column $5$ has the same viewing angle as the occupancy snapshot in column $4$. The target objects retrieved by natural language are highlighted with orange in columns $4$-$5$. 
From Fig.~\ref{fig:vis_supp}, we observe that our VEON succeeds in recognizing open-vocabulary classes such as construction worker, bus, and truck. This proves the efficacy of our model in open-vocabulary 3D occupancy prediction in the wild. 

\section{Potential Societal Impact and Limitations}

\subsection{Potential Societal Impact}

Our VEON aims to predict open-vocabulary 3D occupancy, which is a central task in autonomous driving. Such perception around the ego car is not related to privacy-related issues. However, imperfect occupancy prediction results may lead to failure in subsequent planning and control, causing traffic accidents and casualties. We believe that our work makes a solid step towards robust and practical open-vocabulary 3D occupancy prediction, and can inspire further advancements in this essential module for autonomous driving.

\subsection{Limitations}

One major limitation of VEON is that its performance is hindered by the frozen foundation models. For instance, VEON does not perform well on superclasses such as ``other flat'' (see Tab.~\ref{tab:base_main} in the manuscript). This can attributed to the failure of the open-vocabulary segmentor SAN~\cite{cvpr23-san} in recognizing stuff within ``other flats'', including subclasses such as traffic island, river, etc. And the performance of SAN relies on the pretrained CLIP backbone~\cite{cvpr23-san}. Since transferring knowledge from pretrained foundation models is a prevailing trend, we may consider leveraging more powerful Vision-Language Models (VLMs) in the future. These VLMs, \eg MiniGPT-4~\cite{iclr23-minigpt4}, LLaVa~\cite{nips23-llava}, and Qwen-VL~\cite{arxiv23-qwen}, possess strong vision-language comprehension and reasoning capabilities, which may benefit open-vocabulary 3D occupancy prediction.

% ---- Bibliography ----
%
% BibTeX users should specify bibliography style 'splncs04'.
% References will then be sorted and formatted in the correct style.
%

\end{document}